\documentclass[acmtog,screen]{acmart}
\AtBeginDocument{%
  }

\setcopyright{cc}
\setcctype{by-nc-nd}
\acmJournal{TOG}
\acmYear{2026} \acmVolume{45} \acmNumber{4} \acmArticle{94}
\acmMonth{7} \acmDOI{10.1145/3811285}

\usepackage{amsmath,amsfonts,bm}

\def\eqref#1{equation~\ref{#1}}

\def\1{\bm{1}}

\def\vb{{\bm{b}}}

\DeclareMathAlphabet{\mathsfit}{\encodingdefault}{\sfdefault}{m}{sl}
\SetMathAlphabet{\mathsfit}{bold}{\encodingdefault}{\sfdefault}{bx}{n}

\usepackage{color}
\usepackage{multirow}
\usepackage{overpic}
\usepackage{wrapfig}
\usepackage{colortbl}
\usepackage{xcolor}
\usepackage{caption}

\usepackage{subcaption}
\usepackage{enumerate}
\usepackage{algorithm}
\usepackage{algpseudocode}  

\usepackage{url}
\usepackage{color,xcolor,hyphenat,balance,booktabs}
\usepackage{overpic,wrapfig}
\usepackage{diagbox}
\usepackage{physics,amsthm}
\newcommand{\name}{\textsl{ComboStoc}}

\newcommand{\XR}[1]{{#1}}
\newcommand{\FF}[1]{{#1}}
\newcommand{\FFF}[1]{{#1}}
\newcommand{\under}[1]{\underline{\textbf{#1}}}

\def \eg {{\emph{e.g}.\thinspace}, }
\def \ie {{\emph{i.e}.\thinspace}, }

\acmJournal{TOG}

\begin{document}

\title{ComboStoc: Combinatorial Stochasticity for Diffusion Generative Models}

\author{Rui Xu}
\affiliation{\institution{The University of Hong Kong} 
\authornote{Work partially done at MSRA. † Corresponding authors.}
\country{China}}
\email{ruixu1999@connect.hku.hk}

\author{Jiepeng Wang}
\affiliation{\institution{The University of Hong Kong} 
\country{China}}
\email{jiepeng@connect.hku.hk}

\author{Hao Pan}
\authornotemark[2]
\affiliation{\institution{Tsinghua University} 
\country{China}}
\email{haopan@tsinghua.edu.cn}

\author{Yang Liu}
\affiliation{\institution{Microsoft Research Asia} 
\country{China}}
\email{yangliu@microsoft.com}

\author{Xin Tong}
\affiliation{  \institution{Microsoft Research Asia}
\country{China}}
\email{xtong.gfx@gmail.com}

\author{Shiqing Xin}
\affiliation{\institution{Shandong University} 
\country{China}}
\email{xinshiqing@sdu.edu.cn}

\author{Changhe Tu}
\affiliation{\institution{Shandong University} 
\country{China}}
\email{chtu@sdu.edu.cn}

\author{Taku Komura}
\affiliation{\institution{The University of Hong Kong} 
\country{China}}
\email{taku@cs.hku.hk}

\author{Wenping Wang}
\authornotemark[2]
\affiliation{  \institution{Texas A\&M University}
\country{USA}}
\email{wenping@tamu.edu}

\begin{abstract}
In this paper, we study an under-explored but important factor of diffusion generative models, i.e., the combinatorial complexity. 
Data samples are generally high-dimensional, and for various structured generation tasks, additional attributes are combined to associate with data samples.
\FF{We show that the space spanned by the combination of dimensions and attributes can be insufficiently covered by existing training schemes of diffusion generative models, potentially limiting test time performance.}
We present a simple fix to this problem by constructing stochastic processes that fully exploit the combinatorial structures, hence the name \name.
Using this simple strategy, we show that network training is significantly accelerated across diverse data modalities, including images and 3D structured shapes.
Moreover, \name \space enables a new way of test time generation which uses asynchronous time steps for different dimensions and attributes, thus allowing for varying degrees of control over them.
Our code is available at: \url{https://github.com/Xrvitd/ComboStoc}.
\end{abstract}


\begin{teaserfigure}
    \centering
    \begin{overpic}[width=\linewidth]{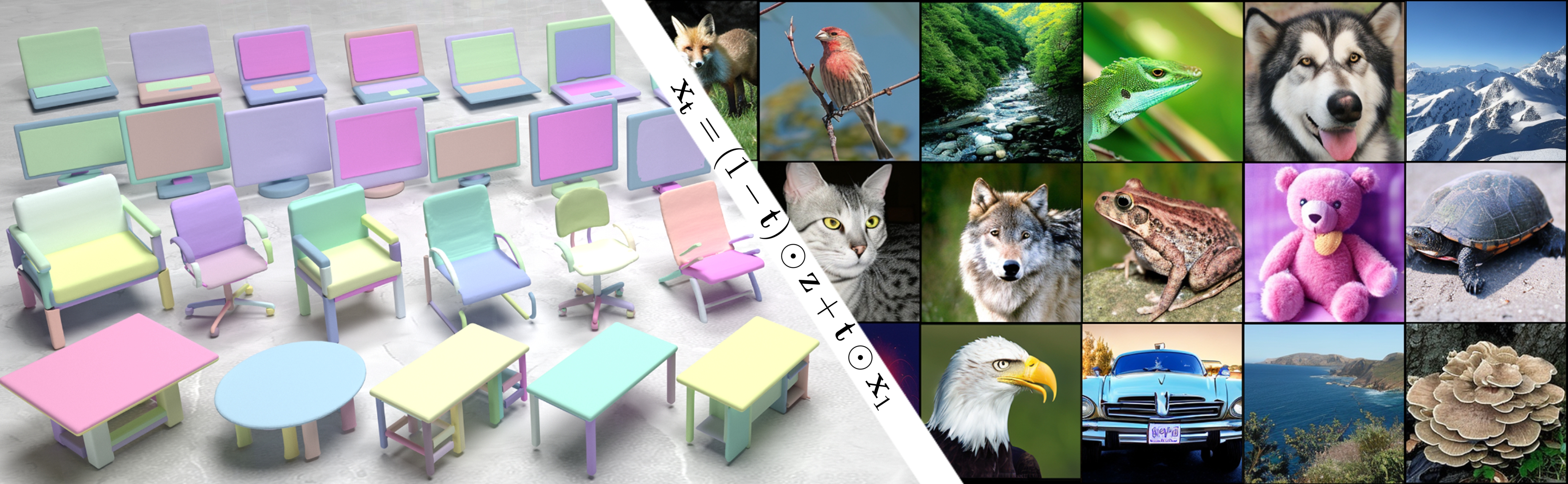}
    \end{overpic}
    \vspace{-6mm}
    \caption{\textbf{\name \space improves diffusion generative models across data modalities of images and structured 3D shapes.} Left: structured 3D shapes where semantic parts are colored randomly. Right: images with consistently lower Frechet Inception Distance (FID) than baseline results. Middle: core idea of \name \space is a simple conversion of the interpolation schedule $t$ of the diffusion models into a tensor of the same shape as the data point $\mathbf{x}_1$ and noise point $\mathbf{z}$, and applying different values within [0, 1] for different dimensions or attributes to fully sample the combinatorial complexity of the dimensions and attributes.}
    \label{fig:teaser}
\end{teaserfigure}

\maketitle


\section{Introduction}
\label{sec:intro}

    
    

Diffusion models are the state-of-the-art generative models across many domains and applications \cite{diffusion_survey}. Diffusion generative models rely heavily on modeling the desired behavior over an extended space of noise corrupted data samples, so that they can cover the target data distributions systematically.
However, the current training schemes generally focus on a single transport path from the source pure noise distribution to the target data distribution \citep{StocIntrp_Albergo2023a,StocIntrp_Albergo2023b,StraightFlow_Liu2022,FlowMatching_Lipman2022}.
\FF{The training therefore can lead to biased sampling density across the space of corrupted samples, where certain regions may be insufficiently covered and, when encountered during stochastic evaluation, produce less accurate behavior.}

To address the mismatch between the training scheme and test-time evaluation, we propose fully sampling the space of combinatorial complexity. The combinatorial nature of this space arises because data samples typically reside in high-dimensional spaces with distinct combinatorial structures.
For instance, the most powerful generative models to date utilize transformers as the network architecture \citep{DiT_2023_ICCV,SiT_Ma2024}. These models treat an image sample as a collection of patch tokens, which are generated in parallel. Furthermore, each patch token is encoded as a high-dimensional vector.
The combination of patches and their feature vectors presents highly complex spaces, over which the diffusion generative models must learn to evolve toward data samples where patches and feature vectors are correlated nontrivially.
In addition, for generative tasks in more structured domains like 3D shapes with semantic parts~\cite{StructureRewriting_Wang23,liu2024part123}, the combinatorial complexity is even more pronounced: each part has numerous attributes encoding different properties like its existence, bounding box and part shape, in addition to the part/patch decomposition and multiple feature channels analogous to images.

\begin{table}[!tp]
\centering
\caption{\FFF{\textbf{Improved convergence over SiT/DiT across iterations.} \name \space can achieve lower FID metrics in the same number of steps  with fewer parameters. 
}}
\vspace{-2mm}
\resizebox{0.85\columnwidth}{!}{
    \begin{tabular}{c|ccc}
    \toprule
Model              & Params(M) & Training Steps & FID  \\ \midrule
DiT-XL             & 675       & 400K           & 19.5 \\
SiT-XL             & 675       & 400K           & 17.2 \\
ComboStoc          & 673       & 400K           & \textbf{15.69}    \\ \midrule
DiT-XL             & 675       & 800K           & 14.3 \\
SiT-XL             & 675       & 800K           & 12.6 \\
ComboStoc          & 673       & 800K           & \textbf{11.41}    \\ \midrule
DiT-XL (cfg=1.5)    & 675       & 7M             & 2.27 \\
SiT-XL (cfg=1.5)    & 675       & 7M             & 2.06 \\
ComboStoc (cfg=1.5) & 673       & \textbf{800K}           & 2.85   \\ 
\XR{ComboStoc (cfg=1.5)} & \XR{673}       & \XR{\textbf{3.1M}}           & \XR{2.14}   \\\bottomrule
\end{tabular}
    }
\label{tab:fids}
\vspace{-5mm}
\end{table}

We sample the spaces of such combinatorial complexity by a simple modification of typical transport plans.
In particular, instead of using a synchronized time schedule for each data sample, we apply asynchronous time steps for each of the patches/parts, attributes and feature vector dimensions, which allows for full sampling of a subspace spanning the various combinations of each pair of source and target data points.

We show that by simply enhancing the training scheme to incorporate the combinatorial sampling, the generative models for images and 3D structured shapes can be significantly improved. 
In particular, for images from ImageNet~\cite{ImageNet}, we obtain systematic FID-50k improvements along different training iterations than baseline SiT~\cite{SiT_Ma2024} and DiT~\cite{DiT_2023_ICCV}~(Tab.~\ref{tab:fids}).
For 3D structured shapes which have stronger combinatorial complexity, we show our training scheme is indispensable for obtaining a working generative model (Fig.~\ref{fig:teaser} left).

In addition to the improved performances, the training scheme exploiting combinatorial stochasticity enables new modes of using the trained generative models.
Specifically, we can now generate different patches/parts/attributes in asynchronous time schedules.
This means that for example we can condition the final sample on flexible partial observations of a reference sample beyond binary masks. 
Instead, for images we can apply graded control across patches and channels.
For structured shapes we can also specify the shapes of some parts only, and let the model generate the remaining parts and attributes.
These new modes of generation have the potential to unify specialized image and shape editing solutions.

\XR{
In summary, we make the following contributions:
\begin{itemize}
    \item We propose \name, an improved diffusion-based training framework that enhances generative modeling by vectorizing the diffusion time steps during training, enabling the model to better capture and reason about structured and combinatorial data.
    \item \name \space performs consistently well across both image and 3D structured shape domains. \name \space significantly accelerates training and achieves lower FID scores on ImageNet~\cite{ImageNet}, while on 3D structured shapes it produces substantially better generation quality on generation tasks.
    \item While maintaining high generation quality, \name \space supports a rich set of timestep–controlled inference applications within shorter training times, including image inpainting, as well as controllable generation and part-level assembly for 3D structured shapes.
\end{itemize}
}




\section{Related Works}
\XR{
\subsection{Image Generation.}
For image generation, a large body of work has focused on improving diffusion training schemes~\cite{go2024addressing,hang2023efficient,wang2024closer,zheng2025beta}, including refined loss weighting and time-step schedules~\citep{Weighting_Hang23}, training acceleration via distillation~\citep{Distillation_Meng23}, and enforcing sampling-path consistency~\citep{Consistency_Song23}. 
Despite these advances, the role of \emph{combinatorial complexity} in diffusion training has received relatively little attention.
A notable exception is \citet{MDT_2023_ICCV}, which attributes the slow convergence of DDPM-based DiT models~\citep{DiT_2023_ICCV} to pixel-wise regression losses that fail to sufficiently emphasize structural correlations across image patches.
To address this issue, \citet{MDT_2023_ICCV} propose a mask-and-diffusion scheme that randomly masks portions of diffused images during training, encouraging the model to learn inter-patch dependencies.
This approach, however, relies on a relatively complex encoder-decoder architecture with additional side-interpolation modules.
In contrast, our method introduces a substantially simpler training scheme that requires only minimal modifications to baseline architectures, yet already yields significant training improvements for SiT~\citep{SiT_Ma2024} models.

Beyond training efficiency, several works explore spatially or temporally varying noise schedules to enable finer-grained control during diffusion inference.
For image editing, SDEdit~\cite{meng2021sdedit} adapts standard diffusion models for stroke-based editing by injecting a global noise level across the entire image, allowing user-provided strokes to be coherently blended into the generated output.
\citet{sahoo2024diffusion} improves likelihood estimation by introducing spatially adaptive noise conditioned on the input signal.
SVNR~\cite{pearl2023svnr} further generalizes this idea with a spatially variant diffusion formulation that initializes denoising directly from the noisy input and assigns each pixel an individual time embedding, enabling more realistic noise modeling and achieving state-of-the-art performance in real-world image denoising.
Soft inpainting is explored in~\citet{levin2025differential} by incorporating carefully designed blending masks into the iterative denoising process.
Related trends also appear in video diffusion models, methods such as~\citet{ruhe2024rollingdiffusionmodels} and~\citet{kim2024fifo} assign stronger noise to later frames to reflect higher temporal uncertainty, while~\citet{chen2024diffusion,song2025history} randomize frame-wise noise strengths during training to simulate diverse prefix-mask conditions.
At inference time, these models again exploit asymmetric noise schedules to better leverage temporal reasoning.
\FF{Concurrently, \citet{huasynchronous} explicitly assign pixel-level timesteps and studies asynchronous denoising for text-to-image alignment, and AR-Diffusion~\cite{sun2025ar} introduces frame-specific timesteps with a scheduler balancing timestep compositions for auto-regressive video generation. Both works share the spirit of per-dimension asynchronous scheduling with our approach, but they focus on different application domains and do not address the training-time combinatorial coverage perspective studied here.}
Different from these approaches, we apply fully unsynchronized noise during training, which allows for graded control in the inference stage of greater flexibility than specific schedules, demonstrating the synergy of efficient training and adaptive inference.

\subsection{Structured 3D Shape Generation.}
While diffusion-based generative models for 3D data have rapidly expanded~\citep{LASDiff_Zheng23,Shape2VecSet_2023_SIG,zhang2024clay,xiang2024structured,zhao2025hunyuan3d}, relatively few works explicitly target \emph{structured} shape generation.
Early efforts such as \citet{StructureNet_Mo19} focus on learning hierarchical shape representations and generating structured variations using a VAE framework.
Building on this line of work, \citet{StructureRewriting_Wang23} propose a rewriting-based model that enables more generalizable cross-category generation.
In contrast to hierarchical representations, we focus on generating \emph{flatly structured} 3D shapes composed of leaf-level semantic parts.
By independently specifying parts and attributes, our model naturally supports a wide range of tasks, including shape completion and part-based assembly.
Previously, these applications were typically addressed using specialized, task-specific solutions~\citep{PartAssembly_Huang20,CompletionPrior_Sung15}.
Our results suggest that such diverse tasks can instead be unified under a single generative framework that explicitly models highly structured data.

Recent part-level generative approaches further highlight the importance of structured 3D representations.
BANG~\cite{zhang2025bang} introduces a diffusion-based framework for part-level decomposition via generative exploded dynamics, producing temporally coherent exploded states that enable controllable and semantically consistent part separation.
X-Part~\cite{yan2025x} proposes a controllable generative model for high-fidelity, structure-coherent part-level decomposition, leveraging bounding-box prompts and point-wise semantic features to produce editable and production-ready 3D assets.
These applications in part-level generation can potentially benefit from a more robust and data-efficient generative model that our approach demonstrates.

\subsection{Diffusion Acceleration and Representation Learning.}
REPA~\cite{yu2024representation} was proposed to accelerate diffusion training by distilling pre-trained, self-supervised visual representations of clean images into intermediate latent representations of noisy inputs, resulting in substantial speedups.
Beyond REPA~\cite{yu2024representation}, several recent studies explore accelerating or stabilizing diffusion and flow-based generative training from other orthogonal perspectives.
DeepFlow~\cite{shin2025deeply} introduces deep velocity supervision with inter-layer velocity alignment, achieving up to an $8\times$ speedup in convergence for flow-based models.
Representation Autoencoders (RAE)~\cite{yu2024representation} replace VAEs with pretrained representation encoders to provide semantically rich, high-dimensional latent spaces, improving reconstruction quality and convergence speed for DiT-style architectures.
Our approach focusing on handling combinatorial complexity is orthogonal to these works, suggesting potential gains from combining these strategies and our approach.
}

\section{Background on Diffusion Models}
\label{sec:background}

The problem of generative modeling aims at capturing the complete distribution of a set of data samples.
Its state-of-the-art solutions include denoising diffusion probabilistic models \citep{DDPM_Ho20}, score-based models \citep{SBGM_Song21} and flow matching \citep{FlowMatching_Lipman2022,StraightFlow_Liu2022}, all of which transform a simple source distribution (\eg the unit normal distribution) into the target distribution following the dynamics specified by variations of stochastic differential equations.
Remarkably, the different formulations can be unified through the framework of stochastic interpolants \citep{StocIntrp_Albergo2023a,StocIntrp_Albergo2023b}.
In particular, the stochastic interpolants framework defines the process of turning data samples into source distributions and vice versa as a simple interpolation between the two distributions, augmented with random perturbations during the processes.

\begin{figure}[tb]
    \centering
    \begin{overpic}[width=\linewidth]{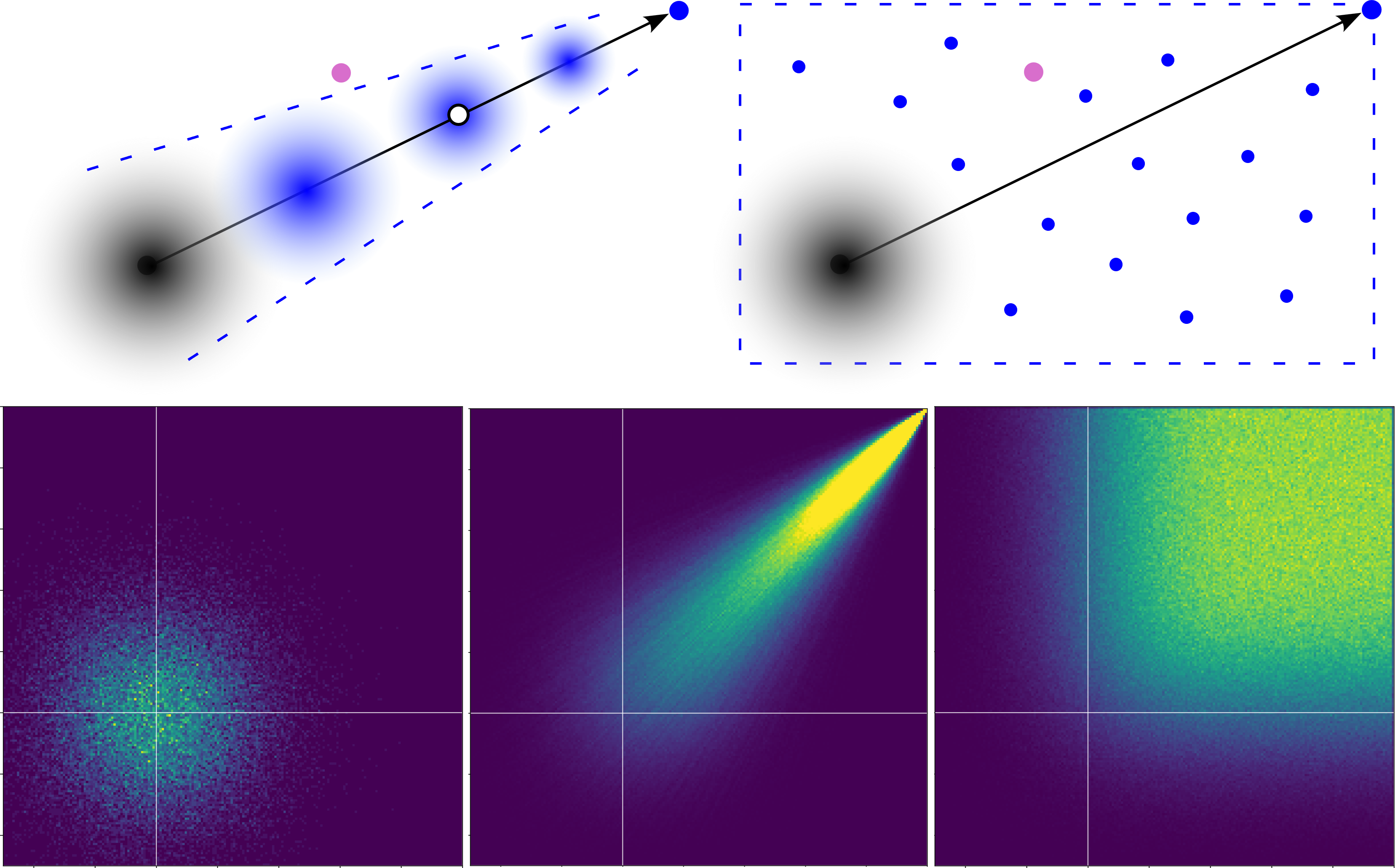}
    \put(2,-3){Noise Distribution}
    \put(38,-3){Flow Matching}
    \put(76,-3){ComboStoc}
    \put(2,37){(a)}
    \put(53.3,37){(b)}
    \put(1,1.4){\textcolor{white}{(c)}}
    \put(34,1.4){\textcolor{white}{(d)}}
    \put(67,1.4){\textcolor{white}{(e)}}
    \put(22,58){ $\mathbf{x}$}
    \put(71.3,58){ $\mathbf{x}$}
    \put(32,50){ $\vb{p}_t$}
    \put(8,42){\textcolor{white}{$\vb{z}$}}
    \put(57,42){\textcolor{white}{$\vb{z}$}}
    \put(48,59){ $\vb{x}_1$}
    \put(98.3,59){ $\vb{x}_1$}
    \end{overpic}
    \vspace{-3mm}
    \caption{\XR{\textbf{\name \space enables better coverage of the generation path space.}} Assuming a single two-dimensional data sample point $\vb{x}_1 = (1.0,1.0)$. 
    \textbf{(a)} The standard linear one-sided interpolant model reduces its density as it approaches individual data samples; the low density regions (like point $\mathbf{x}$) are not well trained and once sampled would produce low-quality predictions. \textbf{(b)} Using \name, for the source noise and target sample points, a whole linear subspace spanned with their connection as the diagonal will be sufficiently sampled, so that there are fewer low-density regions not well trained.
    \textbf{(c,d,e)} We visualize the sampling densities $\rho(\vb{x})$ of the one-sided linear interpolant (d) and \name\ (e) by numerical simulation of a source distribution (c). (d) shows an obvious tendency of shrinking coverage toward the target data point, while (e) has a broader and more uniform coverage.
    }
    \label{fig:idea_illustration}
\end{figure}

\begin{figure*}[tb]
    \centering
    \begin{overpic}[width=0.97\linewidth]{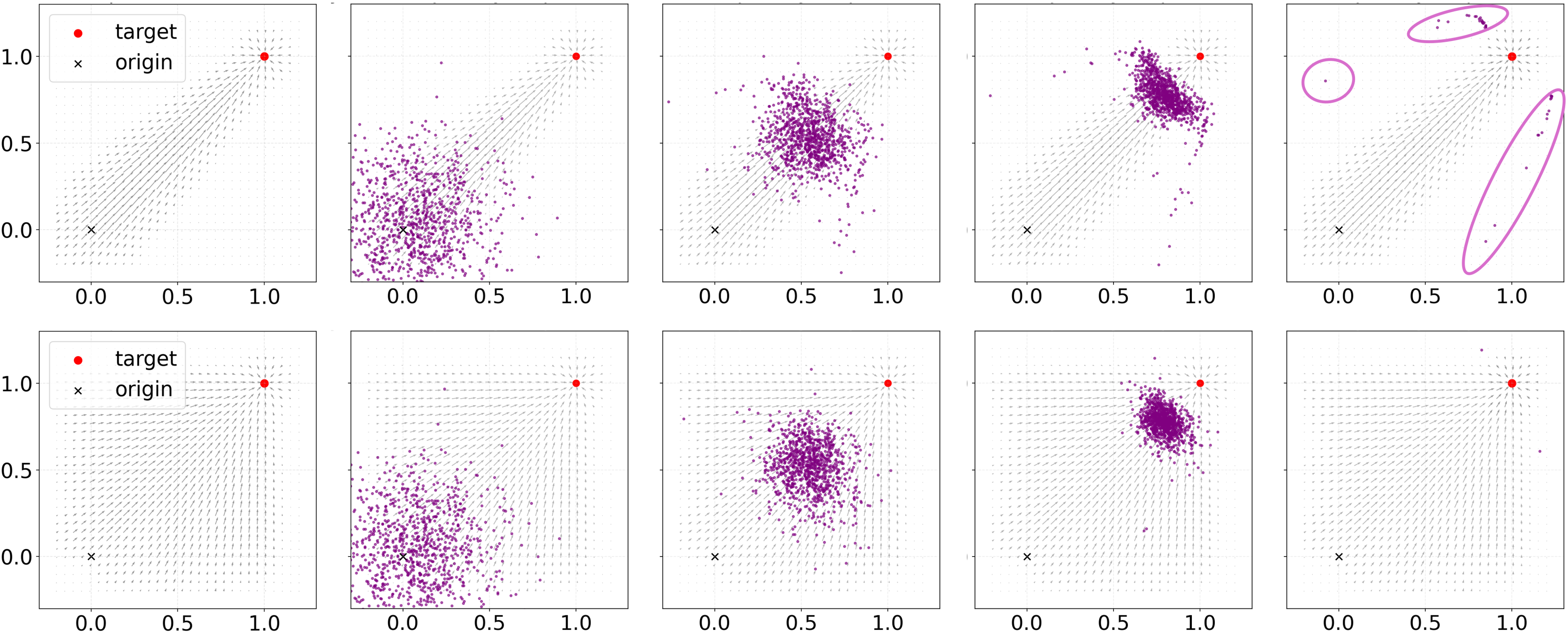}
    \put(-1.7,6){\rotatebox{90}{\textbf{ComboStoc}}}
    \put(-1.7,25){\rotatebox{90}{\textbf{Flow Matching}}}
    \put(7,-1.5){Velocity Field}
    \put(28,-1.5){$t=0.00$}
    \put(48,-1.5){$t=0.50$}
    \put(68,-1.5){$t=0.75$}
    \put(88,-1.5){$t=1.00$}
    \end{overpic}
    \caption{\XR{\textbf{Visualizing the velocity field $u_t(x | x_1)$ and probability density $p_t(x | x_1)$ of the typical one-sided linear interpolant and ComboStoc.} We performed particle simulations on these two velocity fields separately and observed that ComboStoc has fewer outliers. The vector field density follows that of Fig.~\ref{fig:idea_illustration} (d) for flow matching and (e) for ComboStoc, respectively.}
    \FF{This is a 2D particle simulation: 500 particles sampled from an origin-centered (the cross mark) Gaussian distribution are transported according to velocity field discretized by $30{\times}30$ grid. For flow matching, the velocity field is computed by connecting source--target pairs, sampling 100 intermediate points for each pair and averaging velocities from these sampled points. For \name, by definition we use an expanded sampling span for each source--target pair. All particles are integrated using explicit Euler with 100 steps.}
    }
    \label{fig:vfield}
\end{figure*}
We reproduce the formulation of a simple linear one-sided interpolant process below (illustrated in Fig.~\ref{fig:idea_illustration} (a)):
\begin{align}
    \vb{x}_t = (1-t) \vb{z} + t \vb{x}_1,\  t\in [0,1] \label{eq:interp_original}
\end{align}
where $\vb{z} \sim N(0,\vb{1})$ samples the source distribution, $\vb{x}_1 \sim D$ samples the target data distribution, $t\in [0,1]$ is the interpolation schedule. 
A network model $f_{\theta}(\vb{x}_t)$ can be trained to recover the interpolation velocity $\frac{\partial\vb{x}_t}{\partial t} = \vb{x}_1 - \vb{z}$, the target data sample $\vb{x}_1$, or the noise $\vb{z}$ \citep{StocIntrp_Albergo2023b}.
To generate data samples, one starts from random samples $\vb{z}$, follows the velocity field and integrates them numerically to reach the final samples. 
Remarkably, on modeling large scale image datasets like ImageNet~\cite{ImageNet}, a scalable transformer architecture implementing the above process \citep{SiT_Ma2024} shows state-of-the-art performance and outperforms alternative formulations, including DDPM \citep{DDPM_Ho20} implemented via the same network \citep{DiT_2023_ICCV}.

Note that we use the \textbf{linear interpolant model} for its conceptual simplicity. Nonetheless, its practical performance is also strong, being secondary among alternative interpolants according to SiT~\cite{SiT_Ma2024}. 
Importantly, the variety of interpolants all follow an interpolation path $\mathbf{x}_t = \sigma_t\vb{z} + \alpha_t\mathbf{x}_1$, differing only in the specific shape of the path specified by coefficients $\sigma_t,\alpha_t$, which means the problem of undersampling in the path space always exists.

While diffusion generative models are widely recognized for their robustness in modeling diverse distributions by transforming random noise, we note their lack of performance on structured data with insufficient samples (e.g. 3D shapes, 18K in PartNet~\cite{PartNet_Mo19}) and slow convergence on unstructured data with large scale samples (e.g. images, 1.3M in ImageNet~\cite{ImageNet}).
Such deficiencies are shown extensively in Figs.~\ref{fig:2dfid},~\ref{fig:2dexample}, \ref{fig:3dresults_comparison} and Tab.~\ref{tab:3dmetrics} where \texttt{unsync\_none} represents the linear one-sided interpolant model in Eq.~(\ref{eq:interp_original}), presenting failures in structured 3D shape generation and slow convergence in image generation.
\FF{We hypothesize that these difficulties are related to the non-uniform sampling density of the path space, as analyzed next.}

\textbf{Sampling bias.}
We note that although diffusion models are trained to generate from noise, their inputs are more precisely samples on the path space from source to target distributions, which is much larger compared with fixed interpolation paths Eq.~(\ref{eq:interp_original}), and insufficient data samples tend to worsen the problem.
\FF{Specifically, we show that the model by Eq.~(\ref{eq:interp_original}) creates non-uniform sampling density, where regions of the path space farther from the target data points are less densely covered during training.} 

Without loss of generality, suppose that the subspace $\mathcal{R}$ spanned by $\mathbf{z}$ and $\mathbf{x_1}$ has $\mathbf{z}$ as the minimum corner and $\mathbf{x_1}$ as the maximum corner, i.e., $\mathcal{R} = \{ \mathbf{x} | \mathbf{z} \preceq \mathbf{x} \preceq \mathbf{x}_1\}$.
As shown in Fig.~\ref{fig:idea_illustration} (a), we denote an arbitrary point $\mathbf{x}\in \mathcal{R}$, and a moving point $\mathbf{p}_t = (1-t) \mathbf{z} + t \mathbf{x}_1$ along the diagonal connecting $\mathbf{z}$ and $\mathbf{x_1}$.

We denote by $\rho(\mathbf{x})$ the probability density for sampling $\mathbf{x}$, given by integrating all evaluations at $\mathbf{x}$ of Gaussian distributions produced by interpolating the source noise and the target data point, i.e., $G_{\mathbf{p}_t} = N(\mathbf{p}_t; (1-t)^2 \mathbf{1})$ centered at $\mathbf{p}_t$ and with variance scaled by the interpolation coefficient $1-t$. 
Therefore, we have
\begin{equation}
\rho(\mathbf{x}) = \int_{0}^{1} G_{\mathbf{p}_t} (\mathbf{x}) d{t}= \int_0^1 \frac{1}{\sqrt{2\pi}(1-t)} e^{-\frac{\|\mathbf{x}-\mathbf{p}_t\|^2}{2(1-t)^2}} dt.
\end{equation}

Substituting $\mathbf{p}_t$ with the parameterized equation $\mathbf{p}_t = (1-t) \mathbf{z} + t \mathbf{x}_1$, we have
\begin{equation}
    \rho(\mathbf{x}) = \frac{1}{\sqrt{2\pi}}\int_{0}^{1} \frac{1}{1-t} e^{-\frac{\|t(\mathbf{z}-\mathbf{x}_1) + \mathbf{x}-\mathbf{z}\|^2}{2(1-t)^2}} d{t}.
\end{equation}

\XR{
The above integration does not have closed-form solution, so we visualize $\rho$ by numerical integration in Fig.~\ref{fig:idea_illustration} (c,d,e), where we see the clear tendency of shrinking coverage toward data points. Moreover, we find the gradient $\nabla\rho$ is more friendly to work with. In particular, by straightforward calculation we have
\begin{equation}
    (\mathbf{x}_1-\mathbf{x})\cdot \nabla\rho(\mathbf{x}) = \frac{1}{\sqrt{2\pi}}e^{-\frac{\|\mathbf{x}-\mathbf{z}\|^2}{2}} > 0,
\end{equation}
which means that $\nabla\rho(\mathbf{x})$ has a positive projection along the direction $\mathbf{x}_1-\mathbf{x}$. Therefore, we can conclude that the sampling density $\rho(\mathbf{x})$ is not uniform and grows when approaching the target data points.}

\XR{
In Fig.~\ref{fig:idea_illustration}(d) we visualize the sampling density of the standard flow matching formulation, where the shrinking density is clear.
Additionally, in Fig.~\ref{fig:vfield} the first row, we visualize the velocity field $u_t(x \mid x_1)$ and the probability density $p_t(x \mid x_1)$ in a toy example setting, where we observe non-converging outliers in the integrated trajectories.
\FF{This toy example illustrates the potential effects of non-uniform sampling density in flow matching, where regions farther from target data points receive less training coverage.}
}

\textbf{Remark.} The above analysis assumes a source unit normal distribution and a single target data point. When considering the target data distribution as a set of points with variance $\Sigma$, the parameterized Gaussian distribution $G_{\vb{p}_t}$ has variance $(1-t)^2 \mathbf{1} + t^2\Sigma$, which motivates the use of variance-preserving interpolants, such as the cosine/sine interpolant \cite{SiT_Ma2024}.
Nevertheless, the distribution of target dataset may not be large enough to cover the path space well.
For data with many attributes and therefore of high dimensions, the dataset sparsity problem is worsened by the curse of dimensionality. Indeed, we observe that while for images with sufficiently large datasets, mitigating the sampling bias mainly improves convergence, for small scale datasets like structured 3D shapes and sparse images, addressing the sampling bias is essential for training a working generative model (Sec.~\ref{subsec:results_improvement}, Sec.~\ref{subsubsec:imagenet_oneclass}).

To address this biased sampling problem, in the next section we show that by simply desynchronizing the interpolation schedule and sampling all possible combinations of attributes and feature dimensions, the sampling density becomes uniform within the subregions $\mathcal{R}$, thus enabling robust generation in the low-data regime and faster convergence in the rich-data domain.

\section{Combinatorial Stochastic Process}
\label{sec:combo_stoc}

Most interesting data samples are high-dimensional. For example, state-of-the-art generative models encode images as latent patches with both spatial and feature dimensions~\citep{SiT_Ma2024,DiT_2023_ICCV}. 3D shapes structured as part ensembles include even more attributes in addition to spatial and feature dimensions, such as the varying numbers of parts, their bounding boxes, and their positions~\citep{PartNet_Mo19}. Generating such data requires handling more flexible dimensions.

Regardless of the number of dimensions and attributes a data sample has, standard diffusion generative models treat them homogeneously and synchronously. 
For instance, in the case of stochastic interpolants, the generative model is trained on samples distributed according to densities with shrinking coverage along the transport paths connecting the source distribution to each target data sample, as illustrated in Fig.~\ref{fig:idea_illustration}~(a), (d) and analyzed in Sec.~\ref{sec:background}. 
This design leaves the low density regions insufficiently trained, and once they are sampled in test stage by solving stochastic differential equations, the network can produce poor results.

To address the aforementioned problem, we emphasize the combinatorial complexity of individual dimensions and attributes of data samples. Specifically, we purposely sample points with asynchronous diffusion schedules for dimensions and attributes, as illustrated in Fig.~\ref{fig:idea_illustration}~(b). Implementing these asynchronous schedules is relatively straightforward. We transform the interpolation schedule \( t \) from Eq.~(\ref{eq:interp_original}) into a tensor \( \vb{t} \) of the same shape as \( \vb{x} \), using different values independently and uniformly sampled within \([0,1]\) for the dimensions and attributes to obtain sample points:
\begin{equation}
    \vb{x}_{\vb{t}} = (1-\vb{t}) \odot \vb{z} + \vb{t} \odot \vb{x}_1
    \label{eq:interp_insyc}
\end{equation}
where \( \odot \) denotes the elementwise product. 
We note that in contrast to the biased sampling of the standard model (Fig.~\ref{fig:idea_illustration}), the sampling density of \name \space is uniform within the subregions spanned by each pair of source and target data points by construction. 


    
The benefits of using these augmented samples from combinatorial stochasticity are threefold:
\begin{enumerate}[{(1)}]
    \item Ensuring broader network coverage compared to the synchronized schedule, resulting in more robust and higher quality performance during the testing stage.
    \item Encouraging the network to learn the correlations among different dimensions and attributes, as it is trained to synchronize them to reach the final data points.
    \item Enabling more flexible control over the generation process, allowing different dimensions and attributes to be given varying degrees of finalization in the synthesized final result.
\end{enumerate}

\XR{In Fig.~\ref{fig:vfield}, we visualize the velocity field $u_t(x \mid x_1)$ and the probability density $p_t(x \mid x_1)$ in a toy example setting. The velocity field $u_t(x \mid x_1)$ is obtained by summing velocities (Sec.~\ref{app:off-diagonal-compensation}) over all $x_0$-derived spans from Fig.~\ref{fig:idea_illustration} (c,d,e), which does not admit a closed-form expression in this case. Therefore, we additionally perform particle-based simulations to illustrate how noisy samples move toward the data points as the induced velocity field evolves. As shown, our velocity field exhibits broader spatial coverage, resulting in fewer outliers and more concentrated particle trajectories throughout the convergence process.
}


Next, we discuss the detailed adaptations and achieved effects through generative tasks from two different domains, namely images and structured 3D shapes.

\vspace{-2mm}
\subsection{Images}
For image generation, we take on the baseline of SiT~\citep{SiT_Ma2024} which applies highly scalable transformer networks and achieves state-of-the-art performance on ImageNet scale generation.
In particular, a given image is encoded via the VAE encoder from \citet{SD_Rombach22} as a latent image $\vb{x}_1$ of shape $C{\times} H {\times} W$, and the network is trained to predict velocity given the diffused latent image $\vb{x}_t$ (Eq.~(\ref{eq:interp_original})) and optionally conditioned on the image class $c$ and interpolation schedule $t$, \ie $f_{\theta}(\vb{x}_t; c,t) = \vb{x}_1 - \vb{z}$.

\begin{figure}[!htb]
    \centering
    \begin{overpic}[width=.5\linewidth]{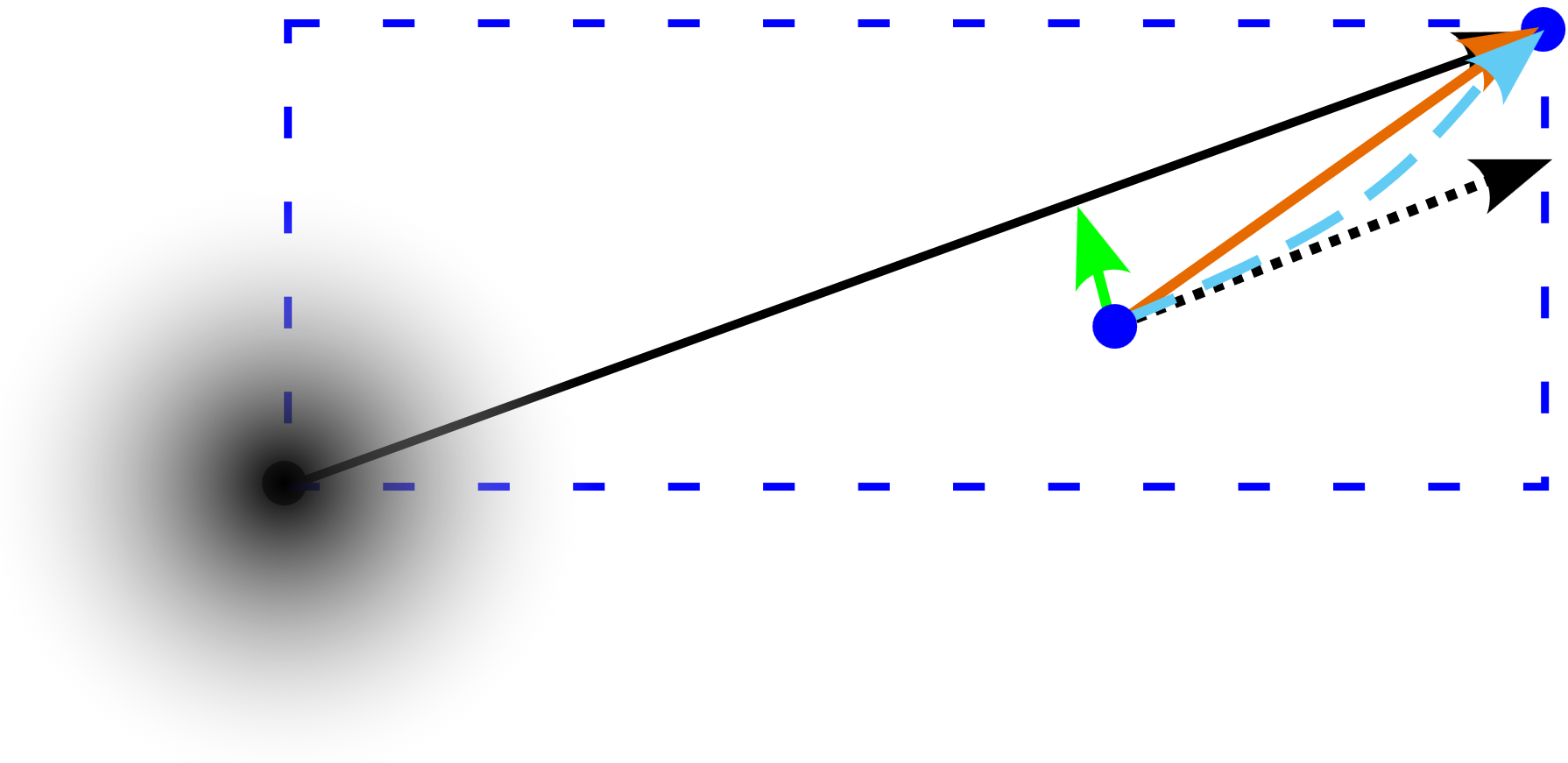}
    \put(12,12){\textcolor{white}{$\vb{z}$}}
    \put(61,27){$\vb{x}_{\vb{t}}$}
    \put(100,50){$\vb{x}_1$}
    \end{overpic}
    \vspace{-6mm}
    \caption{\FFF{\textbf{Illustration of compensation drift.}
When the network is trained to predict velocity $\vb{x}_1 - \vb{z}$ at an off-diagonal sample point $\vb{x}_{\vb{t}}$, a compensation drift ($\vb{v}_{cmpn}$ in \textcolor{green}{green}) can be applied to pull the trajectory back to the diagonal.}}
    \label{fig:cmpn}
\end{figure}

Correspondingly, we make several simple adaptations to implement the \name \space scheme.
First, we construct $\vb{t}$ with the same shape of $C{\times} H {\times} W$, and update the timestep embedding module of SiT to accommodate this change (see Sec.~\ref{subsec:results_improvement} and Fig.~\ref{fig:emb}).
Note that the conditioning on class labels and timesteps are mixed and implemented as modulation operations in SiT~\cite{SiT_Ma2024}, and therefore are not symmetric to the data samples in importance\footnote{The modulation by conditions including class labels and timesteps differ between training and test stages, as during training asynchronous timesteps are used while during testing synchronized timesteps will be used if no graded control is applied (Sec.~\ref{subsec:results_application}). However, the training stage timesteps cover those of test stage as special cases, and thus enhance network generalization. }.
Second, importantly, we note that for velocity prediction, the samples with asynchronous $\vb{t}$ should not predict the original velocity $\vb{x}_1 - \vb{z}$ only; otherwise there will be drift off the target data points during test stage integration, as illustrated by Fig.~\ref{fig:cmpn} the dotted line. 
\XR{Then we provide a formal analysis of this problem.
}

\paragraph{\XR{Proof that the ComboStoc Scheme Defines a Proper Generative Flow Model}}
\label{sec:proof}
Our scheme defines a conditional vector field $u(\vb{x}|\vb{x}_0, \vb{x}_1)$ whenever $\vb{x}\in \textrm{span}(\vb{x}_0, \vb{x}_1)$, where $\textrm{span}(\vb{x}_0, \vb{x}_1)$ is the rectangular subspace spanned by $\vb{x}_0, \vb{x}_1$.
In comparison, FlowMatching \cite{FlowMatching_Lipman2022} defines the conditional vector field $u'(\vb{x}|\vb{x}_0, \vb{x}_1)$ whenever $\vb{x}\in \textrm{diag}(\vb{x}_0, \vb{x}_1)$, where $\textrm{diag}(\vb{x}_0, \vb{x}_1)$ is the line connecting $\vb{x}_0, \vb{x}_1$ and the diagonal of $\textrm{span}(\vb{x}_0, \vb{x}_1)$.
Following FlowMatching, we show that $u(\vb{x}|\vb{x}_0, \vb{x}_1)$ marginalized over $\vb{x}_0, \vb{x}_1$ generates the probability path $p_t$, that connects $\vb{x}_0 \sim p_0 = q(x_0)$ and $\vb{x}_1 \sim p_1 = r(x_1)$.

Precisely, we show that $p_t$ and $u_t$ satisfy the continuity equation:
\begin{align}
    \frac{d}{dt}p_t(x) &= \iint\left(\frac{d}{dt}p_t(x|x_0,x_1)\right)q(x_0)r(x_1)dx_0dx_1 \nonumber\\
    &= - \iint\textrm{div}\left(u(x|x_0,x_1)p_t(x|x_0,x_1)\right)q(x_0)r(x_1)dx_0dx_1 \nonumber\\
    &= -\textrm{div}\left(\iint u(x|x_0,x_1)p_t(x|x_0,x_1)q(x_0)r(x_1)dx_0dx_1\right) \nonumber\\
    &= -\textrm{div}\left(u_t(x)p_t(x)\right), 
\end{align}
where in the first equality we expand the probability into integration over $x_0,x_1$; in the second equality we use the fact that $u(x|x_0,x_1)$ generates $p_t(x|x_0,x_1)$, a point distribution that moves from $x_0$ to $x_1$; in the third equality we switch the order of integration and differentiation based on the regularity of integrands; and in the last equality we simply apply the definition of marginalized vector field, i.e.
\begin{equation}
    u_t(x) = \frac{1}{p_t(x)}\iint u(x|x_0,x_1)p_t(x|x_0,x_1)q(x_0)r(x_1)dx_0dx_1.
\end{equation}

In the above derivations, the major difference from FlowMatching is that we replace the $x_1$-conditioned vector field and probability density from FlowMatching with the $x_0,x_1$-conditioned vector field and probability density, where $u(x|x_0,x_1)$ is a time-invariant vector field that moves the point distribution from $x_0$ to $x_1$ by construction and thus generates $p_t(x|x_0,x_1)$ (Fig.~\ref{fig:idea_illustration}).
\FF{Note that $u(x|x_0,x_1)$ is time-invariant for both FlowMatching and our analysis;
the apparent time dependence in FlowMatching arises only after marginalization over $p(x_0)$.
We also note that the timestep $t$ in the continuity equation above is a \emph{scalar} integration variable: while during training the path space is sampled via vectorized timestep interpolation (Eq.~\ref{eq:interp_insyc}),  the analysis operating in the standard scalar-time framework shows that the resulting marginalized velocity field $u_t(x)$ still generates the correct probability path $p_t$.}
\XR{The issue of off-diagonal drift observed in Fig.~\ref{fig:cmpn} can also be understood from the perspective of test-time integration.
A detailed discussion is provided in Sec.~\ref{app:off-diagonal-compensation}.
}




\subsection{Structured 3D shapes}
We use the generative modeling of structured 3D shapes~\cite{StructureRewriting_Wang23} as a new task to further demonstrate the importance of exploiting combinatorial complexity.
Indeed, structured 3D shapes have even stronger combinatorial complexity than images, as shown in its varying numbers of parts, their positions and bounding boxes, as well as the detailed shape variations for each part.
Precisely, we denote a structured 3D object as a collection of object parts, \ie $\vb{x} = \{\vb{p}_i\}, i\in [L]$, where we set $L=256$ to cover the maximum number of parts in a dataset.
An object part is further encoded as $\vb{p} = (s, \vb{b}, \vb{e})$, where $s\in [0,1]$ indicates the existence of this part, $\vb{b} = (x,y,z,l,w,h)$ denotes the bounding box center $(x,y,z)$ and length $l$, width $w$ and  height $h$, and $\vb{e} \in \mathbb{R}^{512}$ is a latent shape code encoding the part shape in normalized coordinates.
Note that under this representation, a permutation of the part indices does not change the 3D shape, which is quite different from images represented as a feature grid of fixed order and size.

To generate structured 3D shapes with semantic parts, we train a stochastic interpolant model.
In particular, given a structured 3D shape $\vb{x}_1 = \{\vb{p}_i\}$ and its diffused sample $\vb{x}_{\vb{t}}$ (Eq.~(\ref{eq:interp_insyc})), we make the network predict the target data sample directly, \ie $f_{\theta}(\vb{x}_{\vb{t}}; c, \vb{t}) = \vb{x}_1$, where $c$ is the optional class label of the 3D shape.
\FF{We note that for image generation we follow the standard SiT configuration of velocity prediction ($v$-prediction) to isolate the effect of \name, while for structured 3D shapes we adopt $x$-prediction as a more robust design choice for the heterogeneous representation that mixes existence indicators, bounding boxes, and shape codes. As noted by \citet{li2025back}, both velocity and $x$-prediction are viable prediction targets in flow-based generative models.}
Note that here $\vb{t}$ assigns different time steps for all the different attributes and dimensions of each object part.
We validate the generative model for structured 3D shapes by training on the PartNet~\citep{PartNet_Mo19} dataset, as discussed in Sec.~\ref{sec:results}.

\begin{algorithm}[t]
\caption{\XR{Inference with Synchronous and Asynchronous Timesteps}}
\label{alg:combostoc_inference}
\begin{algorithmic}[1]

\Require Trained velocity model $f_{\theta}(\vb{x}_t; c, \vb{t})$; scalar time schedule $\{t_k\}_{k=0}^{K}$ with $t_0 = 0, t_K = 1$; number of steps $K$ (e.g., $K{=}250$).

\Statex
\Statex \hspace{-\algorithmicindent}\textbf{(A) Synchronized Inference}
\State Sample source noise $\vb{z} \sim N(\vb{0}, \vb{1})$.
\State Initialize $\vb{x}^{(0)} \gets \vb{z}$.
\For{$k = 0$ \textbf{to} $K-1$}
    \State $t_k \gets \frac{k}{K}$.
    
    \State Construct synchronized time tensor: 
    \Statex \hspace{\algorithmicindent}$\vb{t}^{(k)} \gets t_k \cdot \vb{1}$ \Comment{Entries $\vb{t}^{(k)}$ share the same scalar $t_k$.}
    \State Predict velocity at this synchronized time:
    \Statex \hspace{\algorithmicindent}$\hat{\vb{v}}^{(k)} \gets f_{\theta}(\vb{x}^{(k)}; c, \vb{t}^{(k)})$ 
    \State Standard SiT-style numerical integration:
    \Statex \hspace{\algorithmicindent}$\vb{x}^{(k+1)} \gets \textsc{SiTStep}\bigl(\vb{x}^{(k)}, \hat{\vb{v}}^{(k)}, t_k, t_{k+1}\bigr)$
    
\EndFor
\State \textbf{return} $\vb{x}^{(K)}$ as the generated sample.

\Statex \hspace{-\algorithmicindent}\hrulefill
\Statex \hspace{-\algorithmicindent}\textbf{(B) Asynchronous Inference with Graded Control}
\setcounter{ALG@line}{0}

\Require Optional observed sample $\vb{x}_1$; mask $\vb{m}\in[0,1]^{\mathrm{shape}(\vb{x})}$  ($m_i$ specifies the degree of preservation for entry $i$, i.e., the initial timestep $\vb{t}^{(0)}=\vb{m}$).

\State Sample source noise $\vb{z} \sim N(\vb{0},\vb{1})$.
\State Initialize the starting point:
\[
\vb{x}^{(0)} = (1-\vb{m})\odot\vb{z} \;+\; \vb{m}\odot\vb{x}_1 ,
\]
\State Initialize the asynchronous timestep tensor:
\[
\vb{t}^{(0)} = \vb{m}, \qquad 
\Delta\vb{t} = \frac{\vb{1}-\vb{m}}{K}.
\]
\Statex Entries with larger $m_i$ start closer to $\vb{x}_1$ and have smaller step size $\Delta\vb{t}$.

\For{$k = 0$ \textbf{to} $K-1$}
    \State Model consumes a fully asynchronous time field $\vb{t}^{(k)}$:
    \Statex \hspace{\algorithmicindent}$\hat{\vb{v}}^{(k)} \gets f_{\theta}(\vb{x}^{(k)}; c, \vb{t}^{(k)})$
    \State Update sample via the standard SiT integrator:
    \Statex \hspace{\algorithmicindent}$
        \vb{x}^{(k+1)} \gets \textsc{SiTStep}
        \bigl(\vb{x}^{(k)}, \hat{\vb{v}}^{(k)}, \vb{t}^{(k)}, \vb{t}^{(k)}+\Delta\vb{t} \bigr)
    $
    \State Update the vectorized timestep:
    \Statex \hspace{\algorithmicindent}$
        \vb{t}^{(k+1)}
        \gets
        \min\!\bigl(\vb{1},\, \vb{t}^{(k)} + \Delta\vb{t}\bigr)
    $
\EndFor
\State \Return $\vb{x}^{(K)}$ as the graded control result.

\end{algorithmic}
\end{algorithm}

\section{Results and Discussion}
\label{sec:results}

In this part, we show that \name \space improves the training convergence of diffusion generative models for both images and structured 3D shapes (Sec.~\ref{subsec:results_improvement}).
We also demonstrate the novel applications enabled by the asynchronous time steps of \name \space (Sec.~\ref{subsec:results_application}).
\FF{We emphasize the distinction between the training and inference contributions of \name: the standard generation quality improvements reported in Sec.~\ref{subsec:results_improvement} (Figs.~\ref{fig:2dfid}--\ref{fig:2dexample}, Tab.~\ref{tab:3dmetrics}) are obtained using \emph{synchronized} timesteps during inference, identical to the baseline SiT inference procedure. Gains arise purely from the asynchronous training scheme. The \emph{asynchronous} inference mode is only employed in Sec.~\ref{subsec:results_application} and Sec.~\ref{sec:step}, where it provides the graded control interface for downstream applications such as soft inpainting and part-level assembly.}

\subsection{Implementation Details}
\label{sec:ImpDetail}


The image generation model is modified from SiT-XL/2, \ie the large model with 28 layers, 1152 hidden dimension, $2\times 2$ patch size, and 16 attention heads.
We trained the model using the default settings of SiT, with AdamW solver and fixed learning rate $10^{-4}$, and batch size 256, on 4 Nvidia H100 gpus.
The training takes 7.5 days for 800K iterations.
Evaluating the models uses the SDE integrator with 250 steps.
The use of classifier-free guidance (CFG) or not is specified at corresponding results.
For comparison with baselines in terms of FID-50K, CFG is not used unless otherwise specified.
In the result gallery Fig.~\ref{fig:teaser}, CFG is used with guidance strength 4.0.

The structured 3D shape generation model uses a network of SiT small model, \ie the model has 12 layers, 384 hidden dimension, $256$ tokens for parts and 6 attention heads.
We trained the model using the AdamW solver with a fixed learning rate of $10^{-4}$ and batch size 16.
We trained the model on 4 Nvidia A100 gpus, which takes 3 days for 1.5K epochs.
Evaluating the models uses iterative sampling with 500 iterations; in each iteration, the predicted part existence is binarized via threshold 0.5 before being diffused back for the next iteration.
Class conditional sampling without CFG is always applied. 
\XR{During typical inferences, we use synchronous timesteps (i.e., the tensorized timesteps have an identical value); 
in applications introduced in Sec.~\ref{subsec:results_application}, we employ asynchronous timesteps for different applications. }

\XR{We provide the pseudocode of the \name\ inference procedure in Alg.~\ref{alg:combostoc_inference}.
For standard generation tasks such as unconditional image synthesis or 3D shape generation, we employ the \emph{synchronized} schedule in Alg.~\ref{alg:combostoc_inference}~(A), where a single scalar timestep is uniformly interpolated into 250 values and broadcast to all dimensions during the denoising process.

For hierarchical control and inpainting applications in Sec.~\ref{subsec:results_application}, we use the \emph{asynchronous} schedule in Alg.~\ref{alg:combostoc_inference}~(B).
In this case, the process does not start from pure noise. Instead, we construct the initial state via $\vb{x}^{(0)} = (1-\vb{m})\odot\vb{z} \;+\; \vb{m}\odot\vb{x}_1,$
where the mask $\vb{m} \in [0,1]^{\mathrm{shape}(\vb{x})}$ encodes the desired degree of preservation for each dimension.
Each entry of $\vb{m}$ simultaneously determines the starting point $\vb{x}^{(0)}$ and its corresponding initial timestep $t^{(0)}_i = m_i$.
The remaining trajectory from $t^{(0)}_i$ to $1$ is then uniformly interpolated into 250 steps.

As a result, different dimensions follow different effective timestep schedules, larger $m_i$ values evolve more slowly, while smaller ones evolve more freely, yet all dimensions complete their evolution within the same 250 integration steps.
This unified framework enables flexible graded control and smooth spatially varying inpainting within a single generative process.
We present the results of synchronous inference in Sec.~\ref{subsec:results_improvement}, explore various applications enabled by asynchronous inference in Sec.~\ref{subsec:results_application}, and report ablation studies comparing the uniform step size and uniform step number strategies in Sec.~\ref{sec:step}.
}


\begin{table}[tb!]
    \caption{\textbf{Enumerating configurations of different combinatorial complexities}, for image domain generation (a) and structured 3D shape generation (b). } 
    \vspace{-4mm}
    \begin{subtable}{.5\linewidth}
      \centering
        \caption{Image domain}
        \vspace{-1.3mm}
        \resizebox{!}{.14\textwidth}{
        \begin{tabular}{c|cc}
        \toprule
        \backslashbox{\textbf{Feature}}{\textbf{Spatial}} & \textbf{w/o patch} & \textbf{w/ patch}\\ \midrule
        single code & \texttt{unsync\_none} & \texttt{unsync\_patch} \\
        feature vector & \texttt{unsync\_vec} & \texttt{unsync\_all} \\ \bottomrule
        \end{tabular}
        }
    \end{subtable}%
    \begin{subtable}{.5\linewidth}
      \centering
        \caption{Structured 3D shape}
        \vspace{-2mm}
        \resizebox{!}{.16\textwidth}{
        \begin{tabular}{c|cc}
        \toprule
        \backslashbox{\textbf{Feature}}{\textbf{Spatial}} & \textbf{w/o part} & \textbf{w/ part}\\ \midrule
        single code &\texttt{unsync\_none}       &  \texttt{unsync\_part}      \\
        attribute &\texttt{unsync\_att}     &  \texttt{unsync\_att\_part} \\
        feature vector &\texttt{unsync\_vec}        &  \texttt{unsync\_all}      \\ \bottomrule
        \end{tabular}
        }
    \end{subtable} 
    \label{tab:all_configs}
\end{table}

\subsection{Improved Training of Diffusion Models}
\label{subsec:results_improvement}

We explore the combinatorial complexities for both images and structured 3D shapes, and build corresponding configurations which exploit these complexities to compare with baseline configurations that do not apply asynchronous time schedules. 
We show that the different configurations improve over baseline configurations universally; in addition, the stronger is the combinatorial complexity, the more important our scheme is for training a working model.


\paragraph{Images.}
Following SiT \citep{SiT_Ma2024}, we train on ImageNet \citep{ImageNet} for class-conditioned image generation. 
To fully explore the effects of combinatorial stochasticity, we enumerate four settings with different levels of combinatorial flexibility in diffusing the data samples (see also Tab.~\ref{tab:all_configs}~(a)).
In particular, we use \texttt{unsync\_none}, \texttt{unsync\_patch}, \texttt{unsync\_vec}, and \texttt{unsync\_all} to denote no splitting of time steps, using different time steps for latent image pixels, for latent image channels, and for both image pixels and channels.
We run the different settings on top of the SiT-XL/2 baseline model. 
Considering the difficulty posed by ImageNet~\cite{ImageNet} data size, in each batch we apply the split time steps only to half of the samples and leave the other half unchanged with synchronized time steps, which balances between samples along and off diagonal paths (Fig.~\ref{fig:idea_illustration})\footnote{This batch mixing scheme may be suboptimal. In preliminary tests (Appendix A.1) we found that blending the split timesteps with synchronized ones gives even better results. Searching the optimal scheme is left for future work.}.
\FF{This mixed strategy serves as a form of curriculum learning: for large-scale datasets, fully asynchronous training can increase early optimization difficulty, so retaining a portion of synchronized samples helps stabilize convergence. This is consistent with other diffusion model strategies that employ non-uniform timestep schedules~\citep{li2025back}. In contrast, for the relatively small PartNet dataset (Sec.~\ref{sec:results}), we apply asynchronous schedules to all samples without mixing, as the smaller data scale does not require such stabilization.}
Plots of FID-50K~\citep{FID_NIPS17} with respect to training steps are shown in Fig.~\ref{fig:2dfid} (a, b), where classifier-free guidance is not used \XR{and all images are sampled by SDE sampling (velocity field can be used for SDE sampling, via Eq.(4) in the SiT~\cite{SiT_Ma2024} paper).} 

\begin{figure}
  \centerline{
  \begin{overpic}[width=\linewidth]{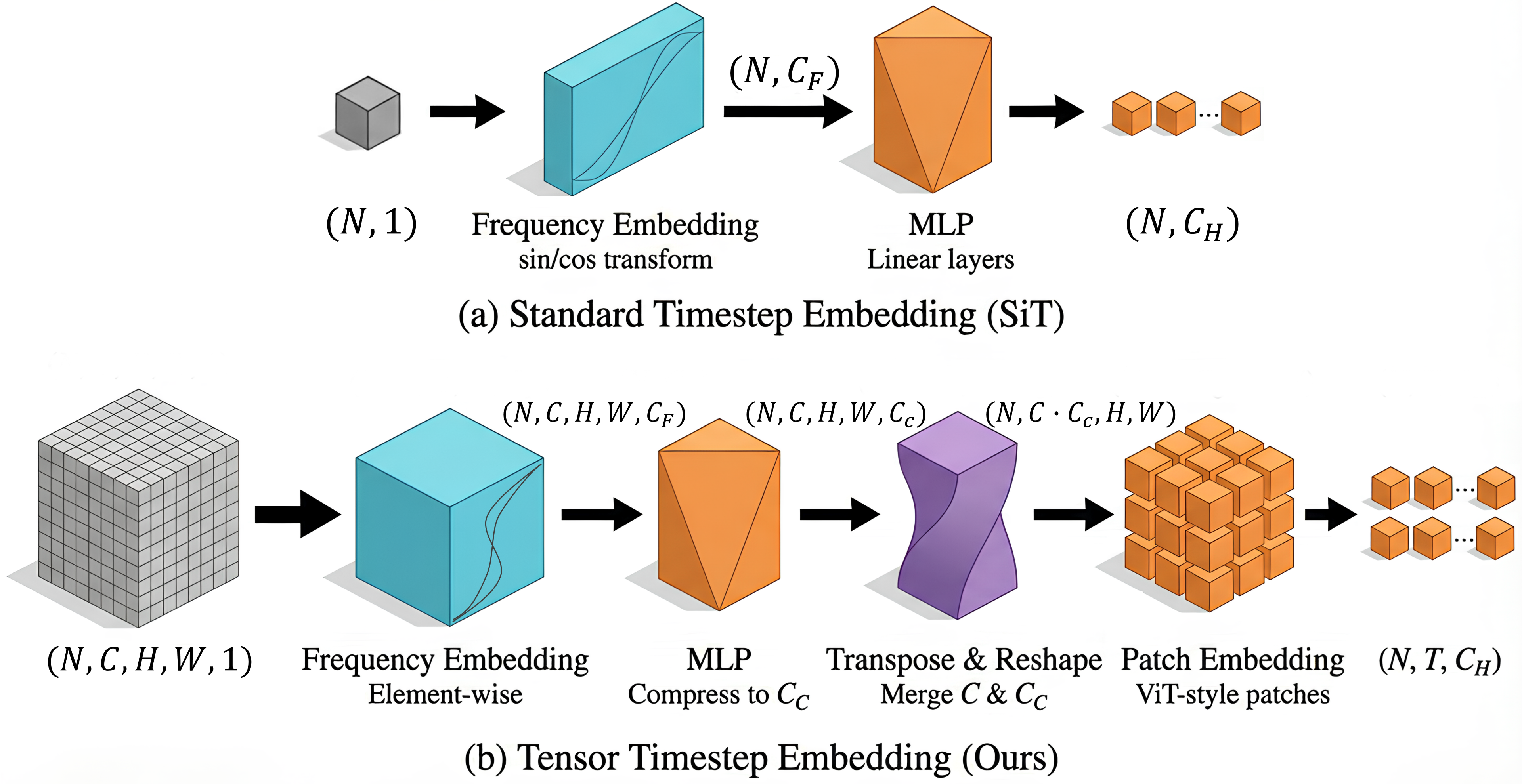}
  \end{overpic}
  }
  \vspace{-4mm}
  \caption{\XR{\textbf{Time embedding module in \name.} 
(a) illustrates the time embedding used in the baseline SiT method, while 
(b) shows the time embedding design adopted by \name.}}
  \label{fig:emb}
\end{figure}

\begin{table}[!tb]
\caption{ \textbf{Time step tensor shapes of different configurations}. \textbf{Left}: images are of shape $(N,C,H,W)$, where $N$ is batch size, $C$ is channel size, $H$ and $W$ are height and width, respectively. The $\vb{t}$ tensors match up with the image tensors through broadcast semantics. \textbf{Right}: structured 3D shapes are of shape $(N,L,[V_s,V_b,V_e])$, where $N$ is batch size, $L$ is the number of shape parts, $[V_s,V_b,V_e]$ is the concatenation of three attributes, \ie $V_s=1$ indicator of existence, $V_b = 6$ bounding box, $V_e = 512$ part shape code; we denote the three attributes collectively as $V$.
$\vb{t}$ tensors match up with the shape tensors through broadcast semantics.}
\vspace{-2mm}
    \begin{subtable}{.5\linewidth}
      \centering
        \caption{Image generation}
        \resizebox{!}{.17\textwidth}{
        \begin{tabular}{cc}
        \toprule
        Setting & $\vb{t}$ \\ \midrule
        \texttt{unsync\_none }      &  $(N,1,1,1)$     \\
        \texttt{unsync\_patch}      &  $(N,1,H,W)$     \\
        \texttt{unsync\_vec }       &  $(N,C,1,1)$     \\
        \texttt{unsync\_all }       &  $(N,C,H,W)$     \\ \bottomrule
        \end{tabular}
        }
    \end{subtable}%
    \begin{subtable}{.5\linewidth}
      \centering
        \caption{Structured 3D shapes}
        \resizebox{!}{.22\textwidth}{
        \begin{tabular}{cc}
        \toprule
        Setting & $\vb{t}$ \\ \midrule
        \texttt{unsync\_none}       &  $(N,1,1)$     \\
        \texttt{unsync\_part}      &  $(N,L,1)$     \\
        \texttt{unsync\_att }     &  $(N,1,[1,1,1])$     \\
        \texttt{unsync\_att\_part}      &  $(N,L,[1,1,1])$     \\
        \texttt{unsync\_vec}        &  $(N,1,V)$    \\
        \texttt{unsync\_all}        &  $(N,L,V)$     \\ \bottomrule
        \end{tabular}
        }
    \end{subtable} 
    \label{tab:all_configs_2}
\end{table}

Given a tensorized timestep $\vb{t}$ of shape $(N,C,H,W)$ that is the same as the latent encoding of input images, we not only encode each of the timesteps for different dimensions as done before via frequency transform \cite{Vaswani17}, but also embed the result feature map of timesteps in the same way as image embedding, \ie the patch-wise embedding originally from ViT~\citep{ViT_21}.
This design ensures that the different dimensions are conditioned on their corresponding timesteps in addition to the shared class label.
In particular, Fig.~\ref{fig:emb}~(a) shows the original SiT timestep embedding module, where $N$ is the batch size, $C_F$ is the length of the sine/cosine frequency embedding, and $C_H$ is the hidden dimension of SiT transformer. Fig.~\ref{fig:emb}~(b) shows the adapted timestep embedding module for \name. $\vb{t}$ is now of shape $(N,C,H,W)$. The first two layers remain the same as the original module, applying to each entry of tensor $\vb{t}$ and producing a compressed timestep encoding of dim $C_C$. Given the result tensor of shape $(N,C,H,W,C_C)$, we further transpose it to combine the channel dimensions and use the same patchwise embedding layer (but different parameters) as SiT and ViT to embed the local patches into vectors of dim $C_H$. Assuming the patch size is $L\times L$, then $T = H\times W/L^2$.
Note that to avoid introducing large embedding layers, we have used $C_C=4$ to encode a timestep scalar, which is significantly smaller than the $C_H=1152$ of SiT.
This can be the reason why \texttt{unsync\_none} performs slightly worse than the baseline SiT, both of which have exactly the same architecture elsewhere.
In Tab.~\ref{tab:all_configs_2} we give the details of split timestep specifications for all configurations, across images and structured 3D shapes.
We rely on the broadcast semantics of Numpy~\cite{numpy} and Pytorch~\cite{paszke2019pytorch} to assign synchronized timesteps to multiple dimensions.

\begin{figure}[!t]
    \centering
    \begin{overpic}[width=\linewidth]{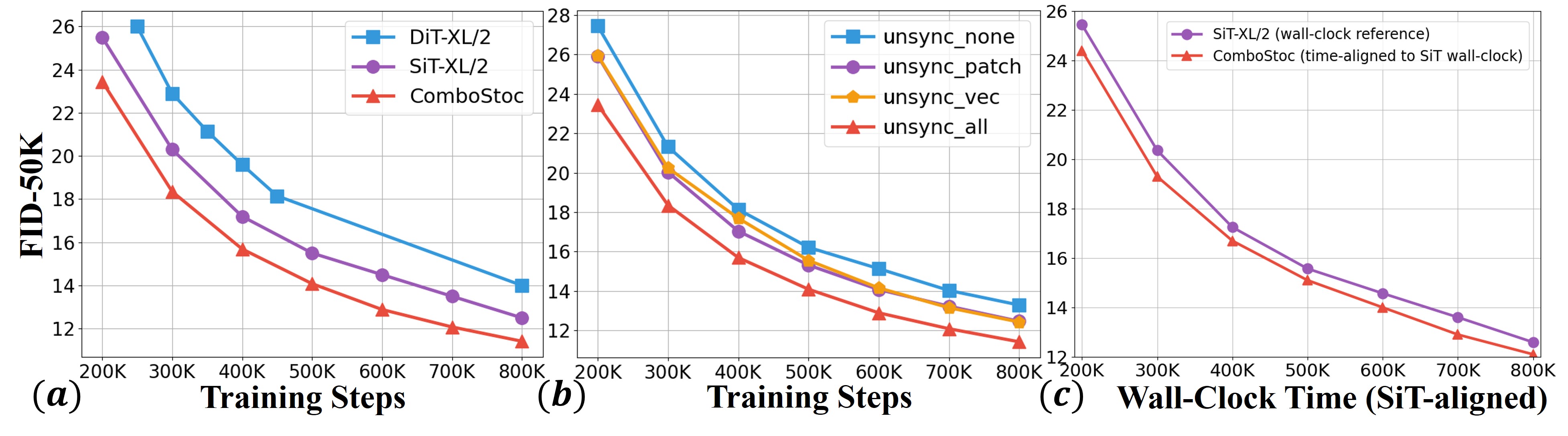}
    \end{overpic}
    \vspace{-7mm}
    \caption{ \textbf{\FF{FID comparison on image generation.}} (a) plots the baseline SiT and our model, as well as DiT for reference; all models are of the scale XL/2 \citep{SiT_Ma2024}. (b) plots the different settings using varying degrees of combinatorial stochasticity. \FF{(c) re-plots (a) with training steps converted to wall-clock time, confirming that \name's improvement holds under a fair time budget. } }
    \label{fig:2dfid}
    \vspace{-1mm}
\end{figure}

\begin{figure*}[!tb]
    \centering
    \begin{overpic}[width=0.94\linewidth]{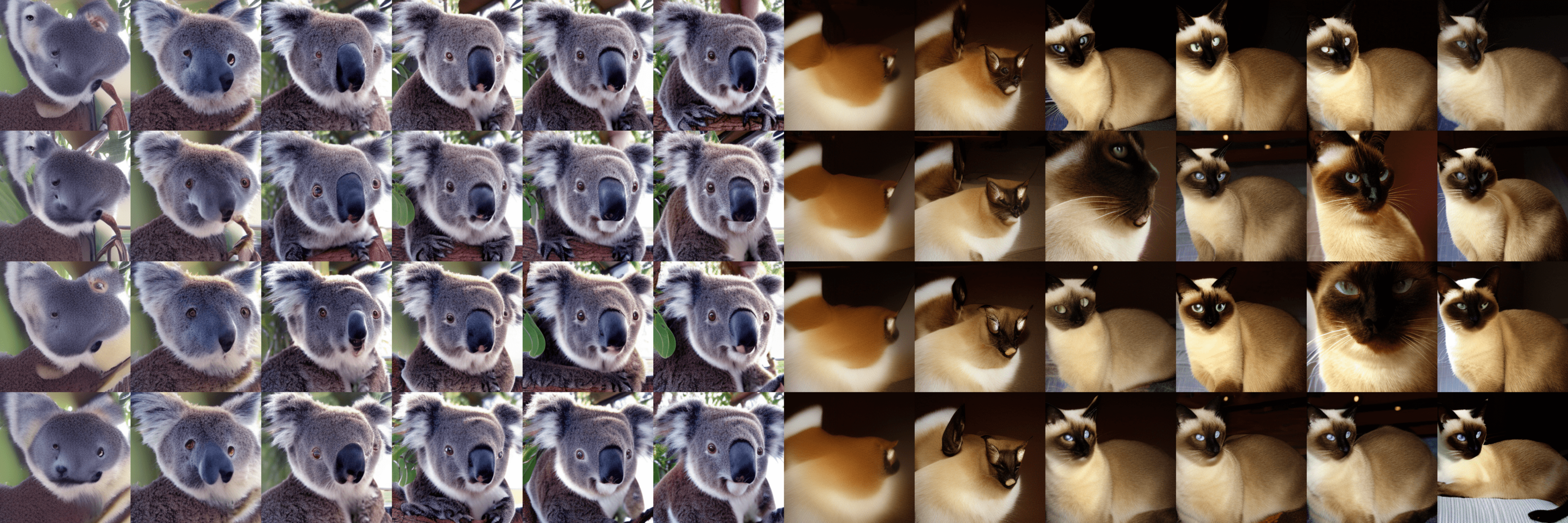}
    \put(-2.3,2){\rotatebox{90}{\small \texttt{all}}}
    \put(-2,11){\rotatebox{90}{\small \texttt{vec}}}
    \put(-2.3,18){\rotatebox{90}{\small \texttt{patch}}}
    \put(-2,27){\rotatebox{90}{\small \texttt{none}}}
    \put(2,-2){\small 50K}
    \put(10,-2){\small 100K}
    \put(18.5,-2){\small 200K}
    \put(27.3,-2){\small 400K}
    \put(35.3,-2){\small 600K}
    \put(43.5,-2){\small 800K}
    \put(52.5,-2){\small 50K}
    \put(60,-2){\small 100K}
    \put(68.7,-2){\small 200K}
    \put(77.2,-2){\small 400K}
    \put(85.3,-2){\small 600K}
    \put(94,-2){\small 800K}
    \end{overpic}
    \caption{
    \textbf{Results of image generation at different training steps}. 
    Settings with stronger combinatorial sampling produce well-structured images earlier; \eg see the koala bear faces and cat eyes. }
    \label{fig:2dexample}
    \vspace{-1mm}
\end{figure*}
As shown by the quantitative results in Fig.~\ref{fig:2dfid} (a), our scheme (using \texttt{unsync\_all}) shows consistent improvement of the baseline SiT model, and significant improvement over the reference DiT model.
\FF{Since \name\ has a slightly slower per-step training speed than SiT due to the tensorized timestep embedding (see Sec.~\ref{sec:app_complexity}), we also provide a wall-clock time comparison in Fig.~\ref{fig:2dfid} (c), which confirms that the improvement holds under a fair time budget.}
Second, as shown in Fig.~\ref{fig:2dfid} (b), the different settings of time step unsynchronization behave differently.
Overall, the finest split by \texttt{unsync\_all} obtains the best performances consistently, followed by \texttt{unsync\_vec} and \texttt{unsync\_patch} which split along feature and spatial dimensions and have almost indistinguishable performances.
The worst performance is obtained by \texttt{unsync\_none}, \ie the setting using no combinatorial stochasticity.
Fig.~\ref{fig:2dexample} visualizes the results of different settings along training steps, where we see better structured images emerge earlier for settings using stronger combinatorial complexity.
\XR{Specifically, at 200K iterations, the \texttt{unsync\_all} setting already produces stable and coherent image structures, whereas the other configurations still exhibit notable fluctuations as training progresses. For instance, under the \texttt{unsync\_vec} and \texttt{unsync\_patch} settings, the facial features of Siamese cats undergo substantial changes between 200K and 600K iterations. Even in the \texttt{unsync\_none} setting, the eyes and facial regions of the Siamese cats remain noticeably less stable compared with the \texttt{unsync\_all} configuration.}
The comparison among these four settings shows that fully utilizing the combinatorial complexity indeed helps network training.
\FF{We reiterate that all FID results in this subsection are obtained with synchronized inference, confirming that the quality gains stem from the asynchronous training scheme alone, without requiring asynchronous inference.}

Due to the smaller timestep embedding module, \texttt{unsync\_none} has slightly worse performance than the baseline SiT.
While it may be possible to align \texttt{unsync\_none} with baseline SiT by introducing more capable embedding layers, \texttt{unsync\_all} already outperforms the baseline with significant margins (Fig.~\ref{fig:2dfid} (a)).
In addition to the result quality, in Sec.~\ref{sec:app_complexity} we provide detailed analysis of the computational complexity of our model in comparison with baseline SiT and DiT models, and find that our model is as efficient as the baselines in actual runtime despite a moderate increase in GFlops.


\begin{table}[tb]
    \centering
    \caption{
    \textbf{Quantitative evaluation of structured shape generation by different settings}. Chair category is used. \underline{\textbf{Best}} scores are marked in bold and underlined; \textbf{second best} scores in bold.
    }
    \vspace{-2mm}
    \resizebox{!}{.13\linewidth}{
    \begin{tabular}{c|cccccc}
\toprule
    & \texttt{none} & \texttt{part} & \texttt{att} & \texttt{att\_part} & \texttt{vec} & \texttt{all} \\ \midrule
    \textbf{FPD$\downarrow$} & 7.99         & 4.71         & 7.47        & \under{3.51}              & 4.62        & \textbf{4.04}        \\
    \textbf{COV$\downarrow$} & 1.32         & 1.03         & 1.83        & \under{0.85}              & 0.97        & \textbf{0.86}        \\
    \textbf{MMD$\downarrow$} & 1.23         & 1.95         & 1.38        & 1.04              & \under{0.63}        & \textbf{0.68}        \\ \bottomrule
    \end{tabular}
    }
    \label{tab:3dmetrics}
\end{table}

\paragraph{Structured 3D shapes.}
We show that for the task of structured 3D shape generation, which has even stronger combinatorial complexity due to the flexible parts and their multiple attributes, our scheme becomes more important to the extent of being indispensable.

We have adopted the pretrained part shape encoding network from \citet{StructureRewriting_Wang23}.
In particular, \citet{StructureRewriting_Wang23} design a point cloud VAE to encode 3D shapes into a sparse set of latent codes, and on top of the latent set, they train another transformer VAE to compress them into a single latent code.
Therefore, each part shape from the PartNet dataset~\citep{PartNet_Mo19} is normalized into unit size and encoded into a single code, which allows us to represent structured 3D shapes as a collection of parts.
The embedding modules for part existence and bounding box follow the same design as timestep embedding.
That is, we first turn each of the scalar dimensions into frequency codes using the sine/cosine embedding, and then embed them into vectors of dim 4 (\textit{cf}. Fig.~\ref{fig:emb}), before finally embedding each of the collective attributes as a whole into vectors of hidden dim 384, through respective FC layers.

For structured 3D shape generation, we identify combinatorial complexity in the following axes: attributes/feature vectors, and spatial parts.
Therefore, we obtain $3{\times} 2 = 6$ settings, \ie \texttt{unsync\_none} and \texttt{unsync\_part} which apply the same or different time schedules to parts respectively, \texttt{unsync\_att} and \texttt{unsync\_att\_part} which use attribute level schedules, and \texttt{unsync\_vec} and \texttt{unsync\_all} which use the most finely divided feature vector level schedules.
See Tab.~\ref{tab:all_configs}~(b) for a summary of the 6 configurations.
Because of the relatively small size of the PartNet dataset (18K shapes in total, mostly in \textit{chair} and \textit{table} classes), we deem it easier to learn and simply apply the corresponding asynchronous time steps to all samples in each batch, in contrast to the mixing scheme of ImageNet training.
We report results at 1.5K epochs, since earlier results cannot be decoded into valid manifold shapes for evaluation in settings like \texttt{unsync\_none}.

\begin{figure*}[!htp]
    \centering
    \vspace{-1mm}
    \begin{overpic}[width=.98\linewidth]{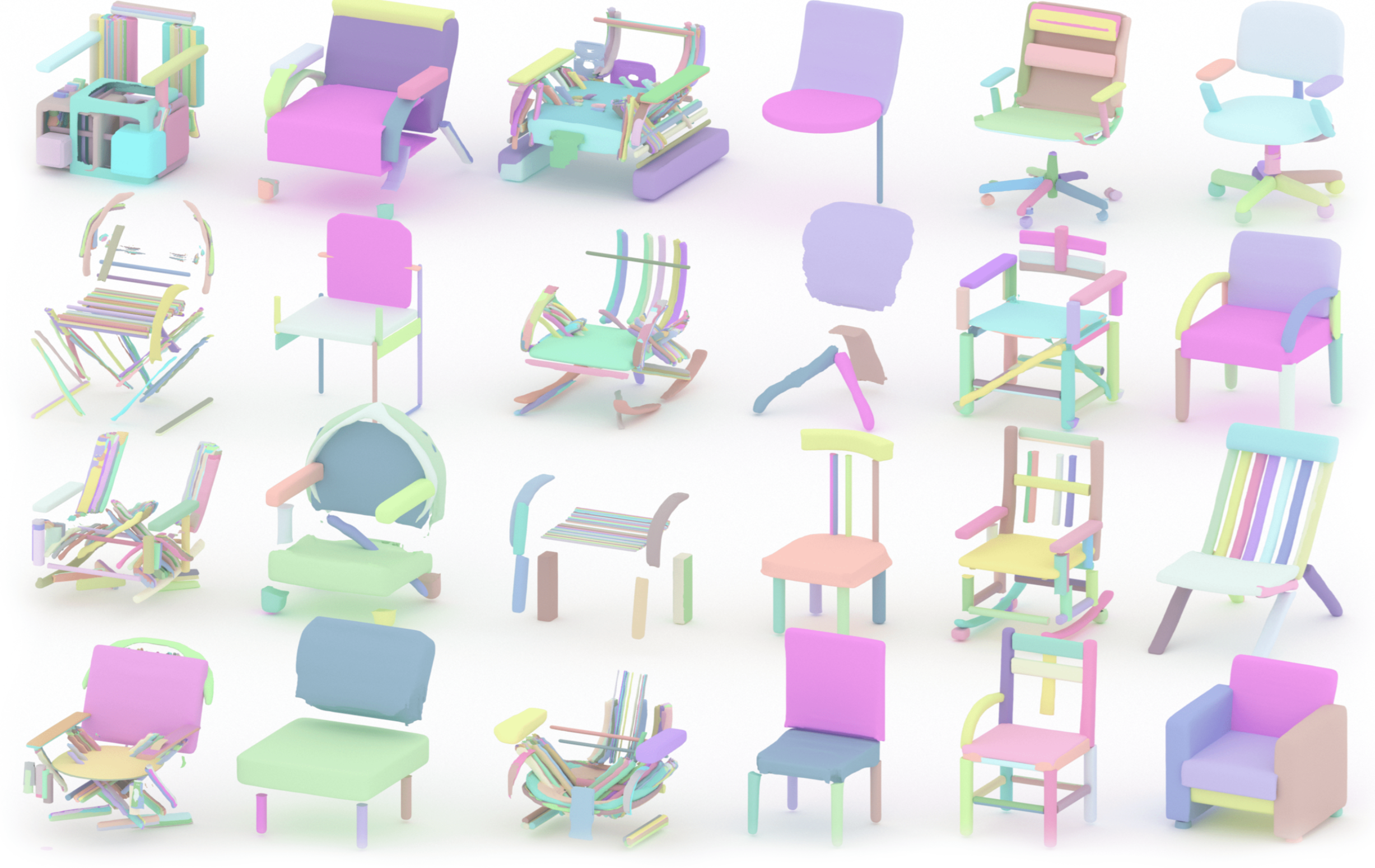}
    \put(2,0){\small \texttt{unsync\_none}}
    \put(18,0){\small \texttt{unsync\_part}}
    \put(36.5,0){\small \texttt{unsync\_att}}
    \put(51.5,0){\small \texttt{unsync\_att\_part}}
    \put(69.5,0){\small \texttt{unsync\_vec}}
    \put(85,0){\small \texttt{unsync\_all}}
    \end{overpic}
    \vspace{-2mm}
    \caption{\textbf{Results of structured shape generation by different settings}. Semantic parts are colored randomly. Settings exploiting stronger combinatorial stochasticity show better results. In comparison, \texttt{unsync\_none}  nearly fails to generate meaningful shapes. }
    \label{fig:3dresults_comparison}
    \vspace{-1mm}
\end{figure*}

\begin{figure}[!tb]
    \centering
    \vspace{2mm}
    \begin{overpic}[width=\linewidth]{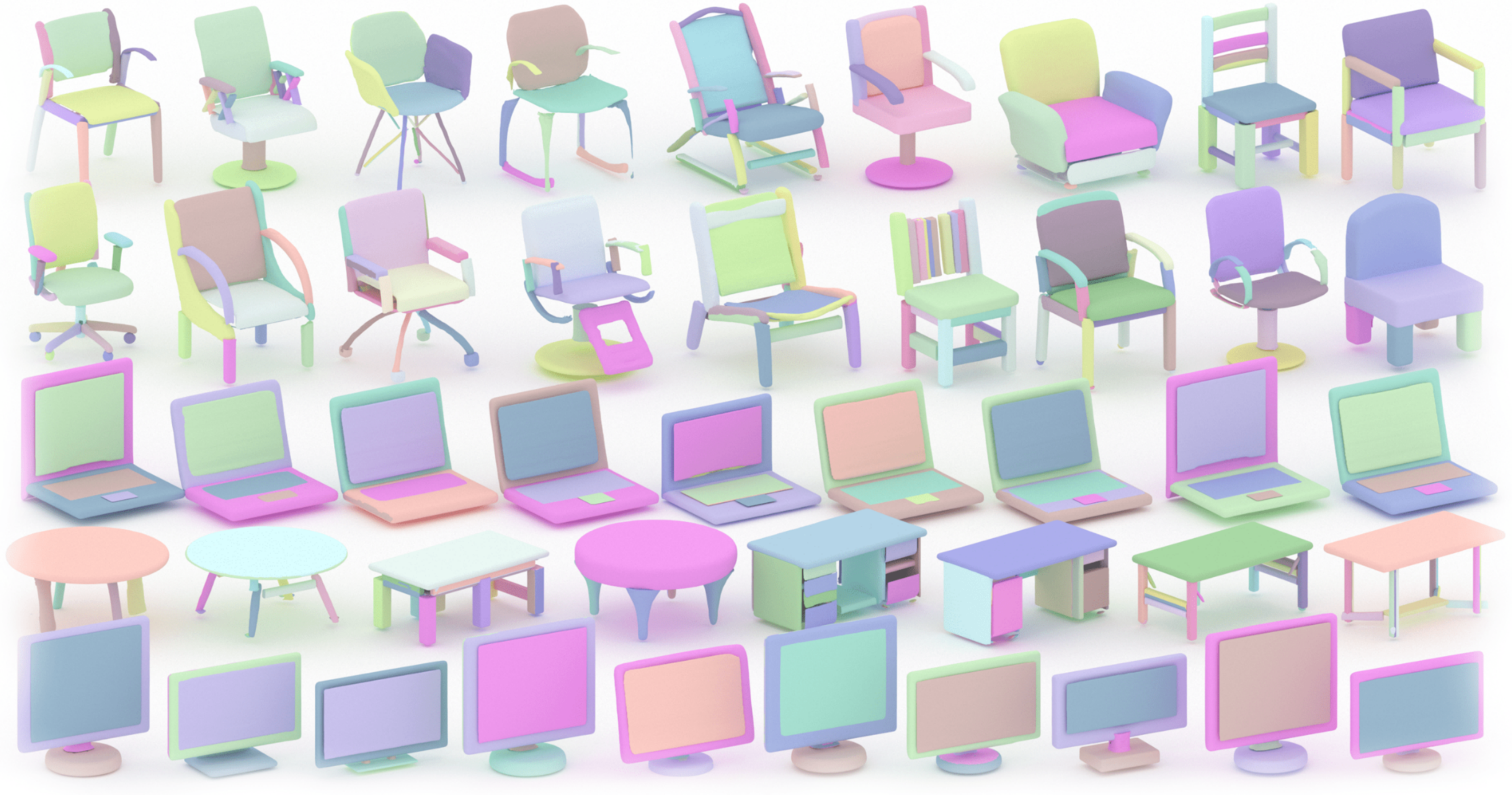}
    \end{overpic}
    \vspace{-7mm}
    \caption{\textbf{Class-conditioned generation of structured 3D shapes.} The classes are: \texttt{chair}, \texttt{laptop}, \texttt{table} and \texttt{display}. }
    \label{fig:3dresults_gallery}
\end{figure}

\begin{figure*}[!t]
    \centering
    \vspace{2mm}
    \begin{overpic}[width=\linewidth]{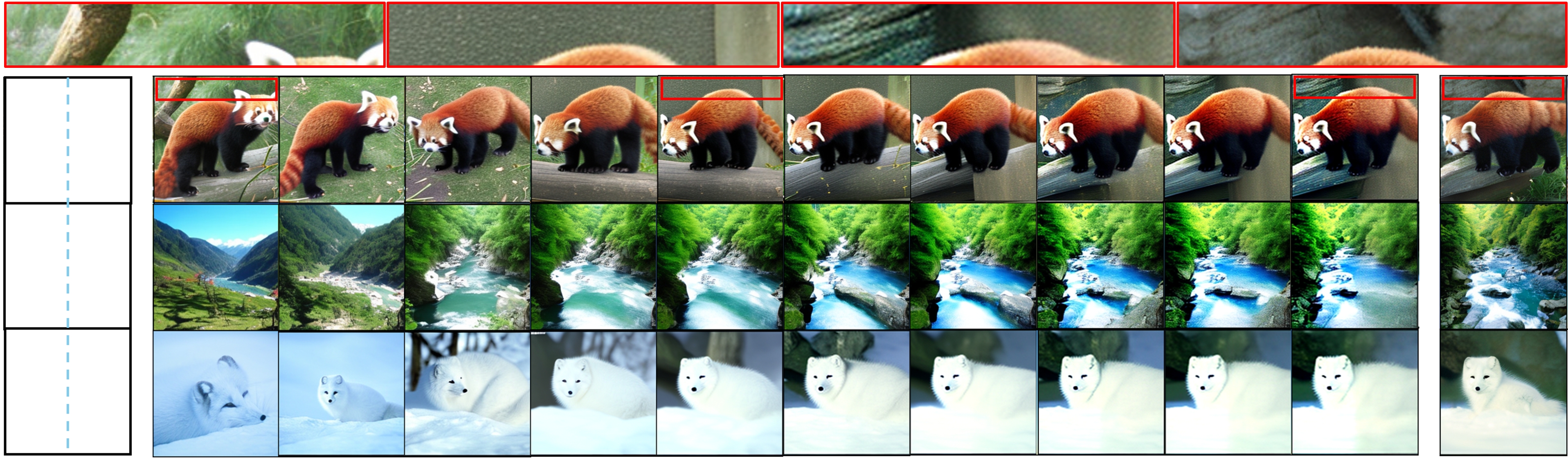}
    \put(1.4,2){\rotatebox{90}{\small $t_0=\lambda$}}
    \put(5.5,2){\rotatebox{90}{\small $t_0=0$}}
    \put(1.4,10){\rotatebox{90}{\small $t_0=\lambda$}}
    \put(5.5,10){\rotatebox{90}{\small $t_0=0$}}
    \put(1.4,18){\rotatebox{90}{\small $t_0=\lambda$}}
    \put(5.5,18){\rotatebox{90}{\small $t_0=0$}}
    \put(11,-2){\small $\lambda=0.0$}
    \put(19,-2){\small $\lambda=0.1$}
    \put(27,-2){\small $\lambda=0.2$}
    \put(35,-2){\small $\lambda=0.3$}
    \put(43,-2){\small $\lambda=0.4$}
    \put(51.3,-2){\small $\lambda=0.5$}
    \put(59.5,-2){\small $\lambda=0.6$}
    \put(67.6,-2){\small $\lambda=0.7$}
    \put(75.6,-2){\small $\lambda=0.8$}
    \put(84.2,-2){\small $\lambda=0.9$}
    \put(95.8,-2){\small $\vb{x}_1$}
    \end{overpic}
    \caption{ \XR{\textbf{Image generation using different weights of preservation.}} Each reference image $\vb{x}_1$ (right) is split into two vertical halves (left), and the left half is given the preservation weights while the right region starts from scratch.}
    \label{fig:2dt0}
\end{figure*}
As shown in Fig.~\ref{fig:3dresults_comparison}, the more combinatorial complexity we exploit, the better the performance of the trained network.
In comparison, the baseline setting without combinatorial stochasticity, \texttt{unsync\_none}, almost entirely fails to produce meaningful shapes.
Moreover, since this task models the highly flexible composition of various numbers of parts, applying the spatial part unsynchronization (Tab.~\ref{tab:all_configs}~(b)) helps obviously, as shown through the three pairs of columns in Fig.~\ref{fig:3dresults_comparison} (\eg \texttt{part} vs \texttt{none}, \texttt{att\_part} vs \texttt{att}, and \texttt{all} vs \texttt{vec}.).

We report quantitative results in Tab.~\ref{tab:3dmetrics} using the \texttt{chair} category. Following~\citet{StructureRewriting_Wang23} we use three metrics, including Frechet Point Distance (FPD) that measures the FID on sampled point clouds, coverage (COV) that measures how well each ground truth sample is covered by the closest generated sample, and minimum matching distance (MMD) that measures how well each generated sample resembles the closest GT sample.
The numerical results again show that the part level combinatorial stochasticity enhances generative performance significantly, and \texttt{unsync\_all} shows the best overall result.

Tab.~\ref{tab:fpd} gives the comparison between our structured 3D shape generation model and two baselines, \ie StructRe~\citep{StructureRewriting_Wang23} and StructureNet~\citep{StructureNet_Mo19}, in terms of FPD, COV and MMD.
Shapes in PartNet~\cite{PartNet_Mo19} are labeled into semantic parts that are organized into trees, \ie coarse parts can be decomposed into fine parts by following the tree.
Exploiting this hierarchical data, the two baselines expand coarse parts into fine parts progressively, which helps constrain the generated shapes toward better regularity.
In comparison, our network does not use this hierarchical information and directly generates the leaf level parts.
Nevertheless, the results by \texttt{unsync\_all} show performances within the baseline results.
Visually, we find our results generally show stronger diversity than the shapes by \citet{StructureRewriting_Wang23} and \citet{StructureNet_Mo19}.
Finally, it is an interesting topic to study how to combine the approaches of hierarchical generation and diffusion generative models, which have differing advantages in aspects of structure regularity and diversity. In Fig.~\ref{fig:3dresults_gallery} we show more random samples generated by the \texttt{unsync\_all} setting.

\begin{table}[!tb]
\centering
\caption{\textbf{Comparison on structured 3D shape generation}. Our results are comparable to the baselines that additionally use the hierarchies of shape parts to constrain generations.
\underline{\textbf{Best}} scores are marked in bold and underlined; \textbf{second best} scores in bold.
}
\vspace{-2mm}
\label{tab:fpd}
\begin{tabular}{ccccc}
\toprule
\textbf{Category}               & \textbf{Method}       & \textbf{FPD$\downarrow$}  & \textbf{COV$ \downarrow$}   & \textbf{MMD$ \downarrow$}  \\ \midrule
\multirow{3}{*}{\texttt{chair}} & StructureNet & 4.67 & 0.89  & \under{0.58} \\
                       & StructRe     & \under{2.63} & \under{0.70}  & \textbf{0.65} \\
                       & Ours         & \textbf{4.04} & \textbf{0.86}  & 0.68 \\ \midrule
\multirow{3}{*}{\texttt{table}} & StructureNet & 6.07 & 1.43  & \textbf{0.55} \\
                       & StructRe     & \under{1.98} & \under{0.66} & \under{0.53} \\
                       & Ours         & \textbf{3.43}    & \textbf{1.20}    & 0.72    \\ \bottomrule
\end{tabular}
\end{table}



\begin{figure*}[!t]
    \centering
    \vspace{-0.5mm}
    \begin{overpic}[width=\linewidth]{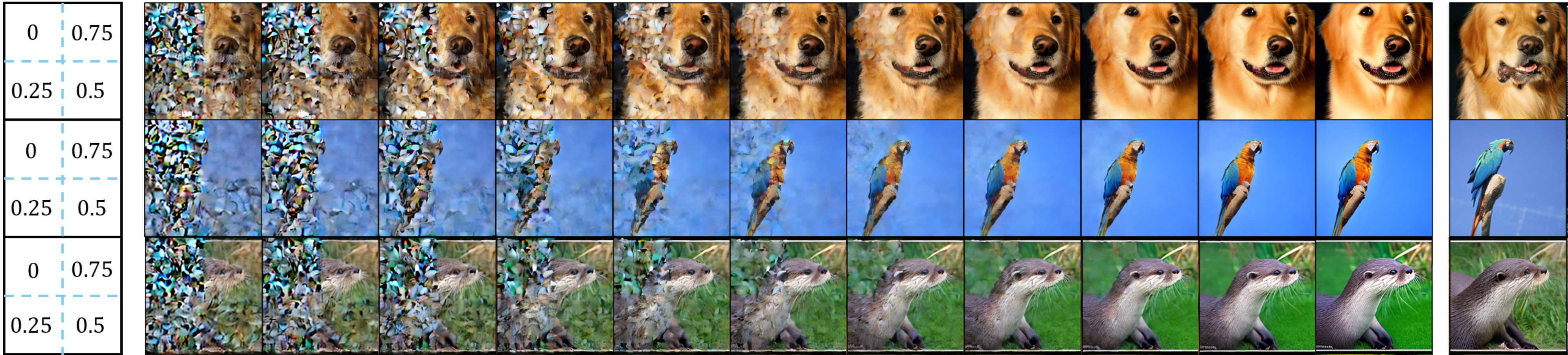}
    \put(3,-2){\small $t_0$ }
    \put(10,-2){\small iter${=}0$}
    \put(19,-2){\small $25$}
    \put(26.5,-2){\small $50$}
    \put(34,-2){\small $75$}
    \put(41,-2){\small $100$}
    \put(48.5,-2){\small $125$}
    \put(55.7,-2){\small $150$}
    \put(63.6,-2){\small $175$}
    \put(70.6,-2){\small $200$}
    \put(79,-2){\small $225$}
    \put(86,-2){\small $250$}
    \put(95,-2){\small $\vb{x}_1$}
    \end{overpic}
    \caption{ \textbf{Image generation with spatially different preservation weights.} As shown in the left column, the four quadrants use $t_0=0, 0.25, 0.5, 0.75$, respectively. The sampling iterations converge to results that preserve the corresponding quadrants from the reference images (right) differently. }
    \label{fig:2dt0iter}
\end{figure*}

\begin{figure}[!tb]
    \centering
    \begin{overpic}[width=\linewidth]{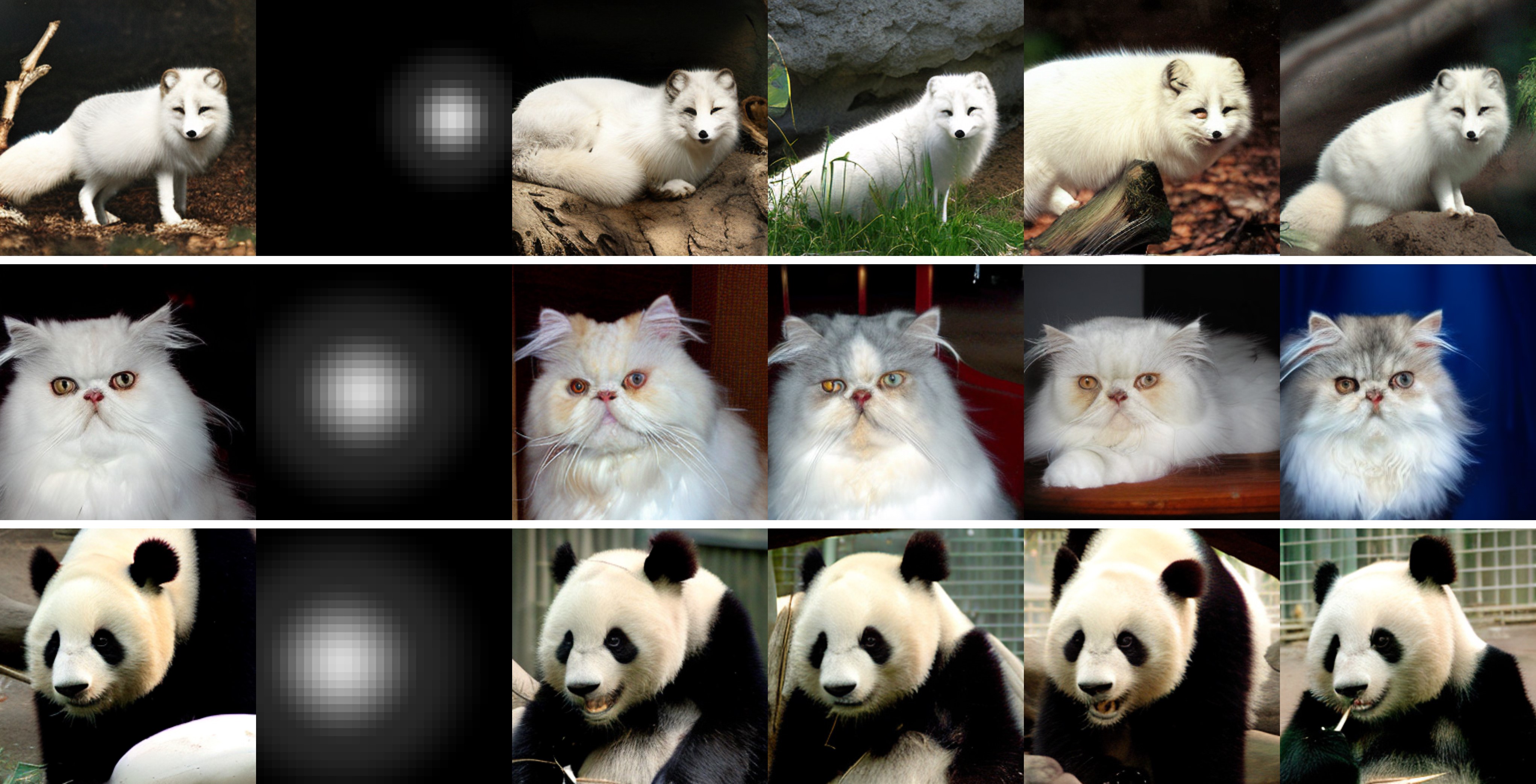}
    \put(4,-3){\small Source}
    \put(20,-3){\small $\vb{t}_0$ Mask}
    \put(56,-3){\small Inpainting Results}
    \end{overpic}
    \vspace{-3mm}
    \caption{ \FF{\textbf{Soft inpainting with spatially continuous $\vb{t}_0$ maps.} The \textbf{first column} shows the source image and the \textbf{second column} shows the $\vb{t}_0$ map (each pixel corresponds to an $8\times 8$ latent patch), which varies smoothly from ${\approx}0.85$ (bright, strongly preserved) to $0$ (black, freely generated). The \textbf{remaining columns} show diverse inpainting results: the model preserves the subject according to the graded mask while coherently generating diverse surroundings. } }
    \label{fig:2dmask}
\end{figure}

\subsection{Applications}
\label{subsec:results_application}

\begin{figure}[!tb]
    \centering
    \begin{overpic}[width=\linewidth]{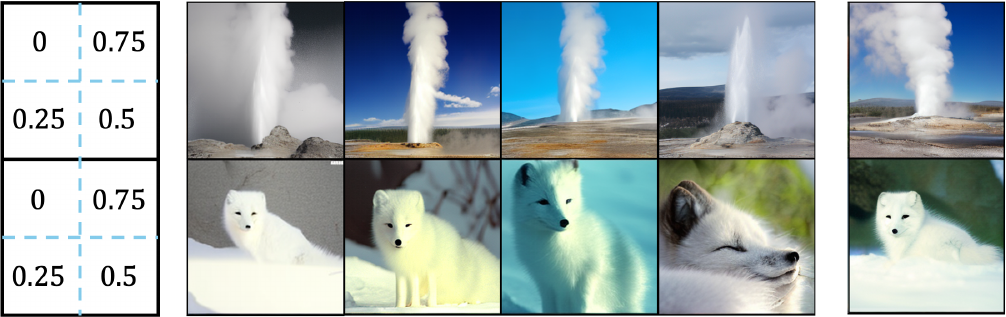}
    \put(8,-2.5){\small $\vb{t}_0$}
    \put(22,-2.5){\small $C=0$}
    \put(38,-2.5){\small $C=1$}
    \put(54.3,-2.5){\small $C=2$}
    \put(70,-2.5){\small $C=3$}
    \put(91,-2.5){\small $\vb{x}_1$}
    \end{overpic}
    \caption{ \textbf{Spatially and channel varying $\vb{t}_0$}. 
    For the spatial dimensions $\vb{t}_0[:,:,i,j]$, the assignment is specified in the left column. For the feature channel dimensions $\vb{t}_0[:,C,:,:]$, the $C$-dim is given 0.5 and the rest given 0. Therefore, we obtain results that correspond to the reference images ($\vb{x}_1$) in complex ways. Notably, earlier channels correspond more to image structures and later channels to image colors.}
    \label{fig:2dvec}
\end{figure}

\begin{figure}[!tb]
    \centering
    \vspace{-1mm}
    \begin{overpic}[width=\linewidth]{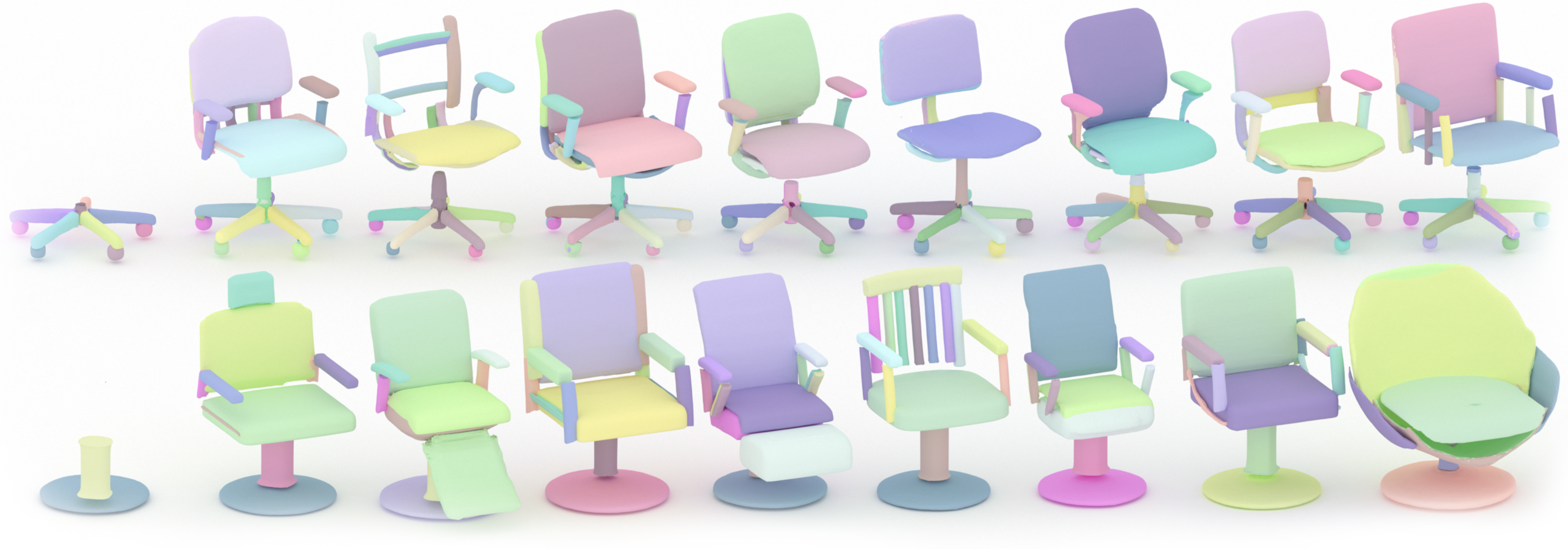}
    \end{overpic}
    \vspace{-8mm}
    \caption{\textbf{Structured shape completion.} Given base parts (left), the network can complete the missing parts conditioned on a shape category name (\texttt{chair} in this example). While the completed parts show great diversity, the given parts are preserved faithfully.}
\label{fig:3dresults_completion}
\end{figure}

\begin{figure}[!tb]
    \centering
    \vspace{-1mm}
    \begin{overpic}[width=\linewidth]{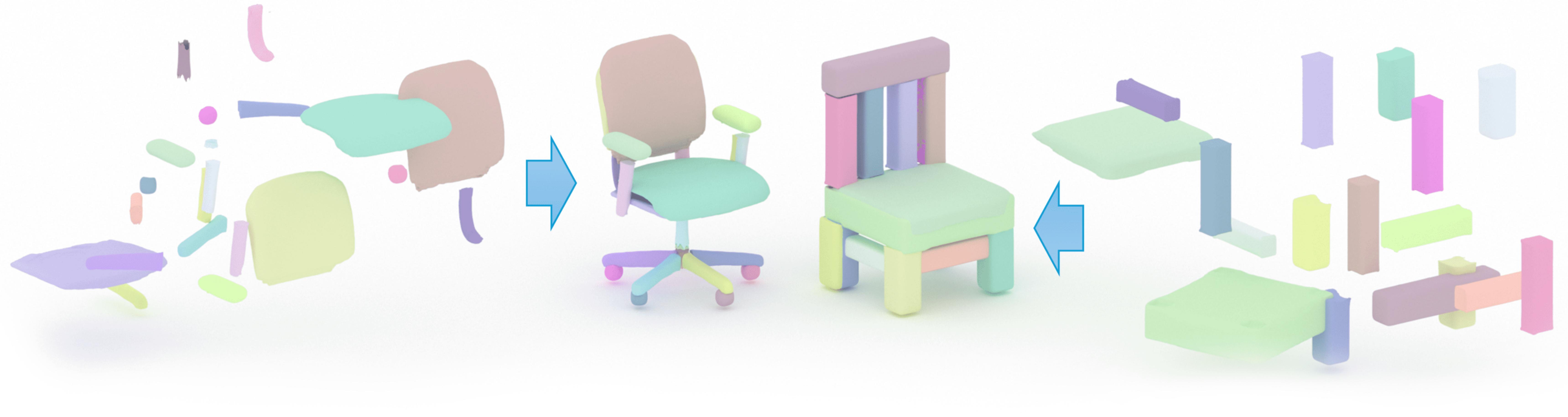}
    \end{overpic}
    \vspace{-10mm}
    \caption{\textbf{Assembly of semantic parts.} Given parts in random positions (left and right), the network assembles them into complete shapes (middle). We solve this part-assembly problem via preserving the attributes of part shapes and scales and only generating the attribute of part positions.}
    \label{fig:3dresults_assembly}
\end{figure}

The asynchronous timesteps for different dimensions and attributes of \name \space enables a novel test-time application, namely the ability to specify different degrees of preservation of a data sample to its dimensions and attributes.
Specifically, given $\vb{t}_0$ specifying the weights in $[0,1]$ to preserve the data of $\vb{x}$, we sample the generative process starting from
 $\vb{x}_0 = (1-\vb{t}_0) \odot \vb{z} + \vb{t}_0 \odot \vb{x}$,
and increase the time steps for individual dimensions and attributes via $\frac{\vb{1} - \vb{t}_0}{N}$ for $N$ steps.
Examples of such asynchronous generative processes are shown in Figs.~\ref{fig:2dt0}, \ref{fig:2dt0iter},~\ref{fig:2dmask},~\ref{fig:2dvec} for images and Figs.~\ref{fig:3dresults_completion}, \ref{fig:3dresults_assembly} for structured 3D shapes.

\paragraph{Images.}
In Fig.~\ref{fig:2dt0} we show that giving different $\vb{t}_0$ to a half of a reference image while leaving the other half to generate from scratch, we can achieve different degrees of preservation of the reference images.
In particular, as the preservation weight increases from $0$ to $1$, the preservation of reference content is strengthened.
\XR{The first example of the red-panda perfectly illustrates the difference in constraint capability caused by the size of $\vb{t}_0$. As $\vb{t}_0$ increases, the area on the left becomes closer and closer to $\vb{x}_1$, especially the background area in the zoom-in window.
We note that at [0.4, 0.7] the weight is good enough to preserve most of the reference content, and if the $\vb{t}_0$ is too big like 0.9, the smooth transition between the two parts will be degraded, since the time steps on the left are too concentrated (only from 0.9 to 1.0).}

In Fig.~\ref{fig:2dt0iter}, we use different preservation weights encoded by $\vb{t}_0$ for the four quadrants of each image, and show intermediate results along the iterative SDE integration process.
From the three examples we can see that stronger weights cause better preservation of reference regions, and the different regions are filled with coherent content despite the spatially varying time schedules.
This mode of controlled generation is novel, compared with the binary inpainting mode proposed for standard diffusion models~\citep{RePaint_Lugmayr22}, where regions of an image are divided into two discrete types, \ie those to preserve and those to generate from scratch.
\FF{Fig.~\ref{fig:2dmask} further demonstrates this capability with smooth, spatially continuous $\vb{t}_0$ maps. Rather than the piecewise-constant weights in Figs.~\ref{fig:2dt0} and~\ref{fig:2dt0iter}, here we use a smoothly varying $\vb{t}_0$ map centered on each subject, with preservation strength ranging continuously from ${\approx}0.85$ at the center to $0$ at the periphery. 
Unlike binary-mask inpainting~\citep{RePaint_Lugmayr22}, \name\ supports such continuous spatial control natively, without any task-specific training.
The generated results show that the model faithfully preserves the central subject while coherently generating diverse surroundings and backgrounds, with natural transitions between preserved and generated regions. 
This smooth, continuous control is a unique advantage of \name's asynchronous timestep formulation.}

We also find that channel-varying $\vb{t}_0$ reveals interesting observations about the different contents of latent image encoding~\citep{SD_Rombach22}.
Fig.~\ref{fig:2dvec} shows another example of image generation where we use varying degrees of data preservation across both spatial dimensions (the four quadrants) and feature channel dimensions.
In particular, we assign spatial preservation weights according to the left column in the figure, and additionally assign 0.5 to the specified channel index $C$ and 0 to other channels, as shown in the middle four columns. 
Interestingly, we see that the different channels of the stable-diffusion VAE latent space \citep{SD_Rombach22} have very different content.
For $C=0$ the first channel, the generated results mostly preserve the spatial structures of the reference images, and the color cues are largely lost.
From $C=1$ to $C=3$, the generated results increasingly preserve the color cues of the reference images but lose more of the structures.
The findings suggest that earlier channels of the VAE latent space emphasize on structures and later ones on image-level color distributions.

\vspace{-2mm}
\paragraph{Structured 3D shapes.}
By controlling different parts and attributes of structured 3D shapes, we can achieve diverse effects, including shape completion and part assembly.
In Fig.~\ref{fig:3dresults_completion}, we fix the bases of chairs by giving them $t_0 = 0.9$, and complete them with meaningful but diverse structures that satisfy the class condition.
The given bases have the chance of being slightly updated to adapt to the completed shapes.
In Fig.~\ref{fig:3dresults_assembly}, we randomly position a set of parts, and let the network arrange them into proper shapes, by giving the part shape codes $\vb{e}$ and bounding box sizes large preservation weights ($t_0 = 0.9$) and making the rest attributes free to be generated.
Here we have considered a simplified setting where the part rotations are given, and leave the more challenging case of rotating shape parts as future work.

\subsection{More Discussions}

\subsubsection{\XR{Minimizing the Off-diagonal Drift}}
\label{app:off-diagonal-compensation}

The problem of off-diagonal drift can be revealed from the perspective of test-time integration. For ease of analysis we assume an ODE is solved to follow the prescribed velocity field to move from the source noise point to the target data point. 

If the integration never stumbles on sample points off diagonal line connecting $\mathbf{z}$ and $\mathbf{x}_1$, the integration starting from a point $\vb{x}_{t_0} = \vb{z}+t_0 (\vb{x}_1-\vb{z})$ and following the velocity field $\mathbf{x}_1 - \mathbf{z}$ would always end at the target point, i.e.,
\begin{equation}
    \mathbf{x}_{t_0} + \int_{t_0}^{1}(\mathbf{x}_1 - \mathbf{z})d{t} = \vb{z}+t_0 (\vb{x}_1-\vb{z}) + (1-t_0)(\mathbf{x}_1 - \mathbf{z}) = \mathbf{x}_1.
\end{equation}

Now suppose the integration stumbles upon an off-diagonal sample point $\mathbf{x}_{\vb{t}_0}$ at a tensorized interpolation schedule $\vb{t}_0$ with different values for its various entries. The integration by following only the velocity $\mathbf{x}_1 - \mathbf{z}$ would miss the target data point $\mathbf{x}_1$ (Fig.~\ref{fig:cmpn}, dotted arrow). Precisely,
\begin{align} 
\mathbf{x}_{t_0} &+ \int_{t_0}^{1}(\mathbf{x}_1 - \mathbf{z})d{t} \\ 
&= \vb{z}+\vb{t}_0\odot (\vb{x}_1-\vb{z}) + (1-t_0)(\mathbf{x}_1 - \mathbf{z}) \nonumber
\\ 
&= \mathbf{x}_1 + (\vb{t}_0-t_0)\odot (\vb{x}_1-\vb{z}), \nonumber
\end{align}
where $t_0 = \min(\vb{t}_0)$ denotes the minimum interpolation schedule across the dimensions of $\vb{t}_0$. Here we assume the test-time setting that for asynchronous timesteps we integrate until the slowest one finishes. 

To address this divergence problem, we propose a velocity compensation approach for mitigation, where the compensated component minimizes the off-diagonal drift by following the negative gradient of a drift potential, as discussed next.



\begin{figure*}[!tb]
    \centering
    \begin{overpic}[width=.95\linewidth]{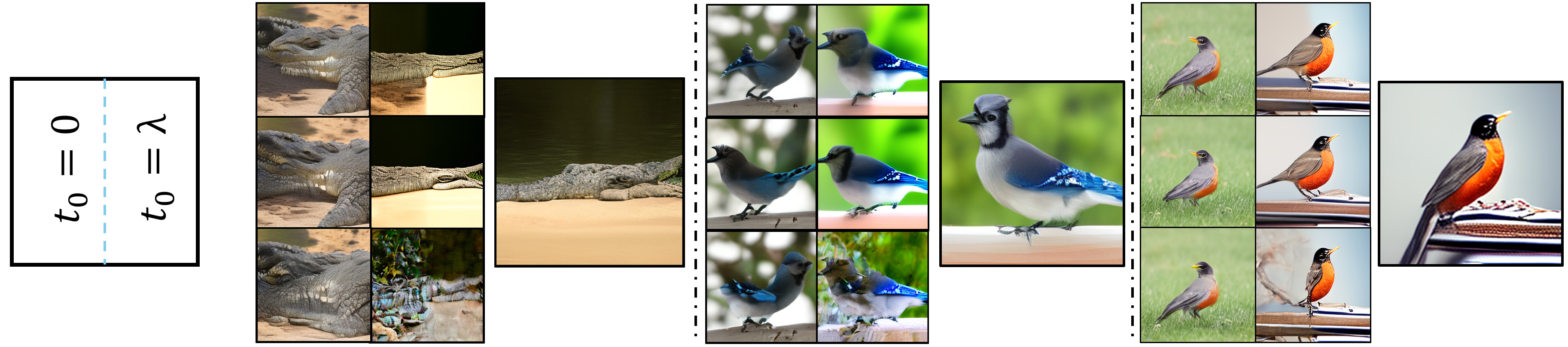}
    \put(14,3.4){{(c)}}
    \put(14,10.2){{(b)}}
    \put(14,17){{(a)}}
    \put(17,-1.5){$\lambda = 0.0$}
    \put(24,-1.5){$\lambda = 0.75$}
    \put(37,4){$\vb{x}_1$}
    \put(46,-1.5){$\lambda = 0.0$}
    \put(53.5,-1.5){$\lambda = 0.75$}
    \put(65,4){$\vb{x}_1$}
    \put(74,-1.5){$\lambda = 0.0$}
    \put(81,-1.5){$\lambda = 0.75$}
    \put(93,4){$\vb{x}_1$}
    \end{overpic}
    \caption{\textbf{Comparison of velocity adaptation methods in graded image generation using different preservation weights.} Each reference image $\vb{x}_1$ is split into two vertical halves, with the right half assigned preservation weights while the left half starts from scratch. Rows (a), (b), and (c) correspond to no compensation, the off-diagonal drift minimization method, and the cone-shaped velocity field method, respectively.}
    \label{fig:drift}
\end{figure*}
\paragraph{Off-diagonal Drift Minimization.}
The off-diagonal offset vector $\mathbf{\delta}(\mathbf{x}_t)$ can be defined as
\begin{equation}
    \mathbf{\delta}(\mathbf{x}_t) = -\mathbf{v}_{cmpn} = \mathbf{x}_{\mathbf{t}} - \mathbf{x}_1 - \frac{(\mathbf{x}_{\mathbf{t}} - \mathbf{x}_1)\cdot(\mathbf{x}_1 - \mathbf{z})}{||\mathbf{x}_1 - \mathbf{z}||^2} (\mathbf{x}_1 - \mathbf{z}).
\end{equation}
To minimize the drift, we can simply follow its negation $\mathbf{v}_{cmpn}$ in addition to the original velocity during integration, which is equivalent to minimizing a drift potential $\Phi\left(\mathbf{\delta}({\mathbf{x}_{\mathbf{t}}})\right) = \frac{1}{2}\|\mathbf{\delta}_{\mathbf{t}}\|^2$ by gradient descent, and promotes the convergence to target data points (Fig.~\ref{fig:cmpn}, dashed arrow).




To verify the effectiveness of our proposed method, we also tried another intuitive velocity adaptation method, where we design a cone-shaped velocity field such that its integration converges to target data points, as discussed next.

\paragraph{Cone-shaped Velocity Field.}
Different from the off-diagonal drift minimization, we can also design a cone-shaped velocity field that generalizes the simple constant velocity $\vb{x}_1-\vb{z}$ to a cone of velocities covering the expanded region $\mathcal{R}$.
In particular, we can use the following velocity 
\begin{equation}
    \vb{v}_{\vb{t}_0} = \frac{\vb{x}_1 - \vb{x}_{\vb{t}_0}}{1-t_0},
\end{equation}
where again $t_0 = \min(\vb{t}_0)$ denotes the minimum interpolation schedule across the dimensions of $\vb{t}_0$.
Note that for synchronized schedule, $\vb{v} = \frac{\vb{x}_1 - \vb{x}_{t_0}}{1-t_0} = \frac{\vb{x}_1 - \vb{z} - t_0 (\vb{x}_1-\vb{z})}{1-t_0} = \vb{x}_1 - \vb{z}$, i.e., the original velocity is a special case of this velocity field.
To see why this is a cone shaped velocity field, note that for a timestep $\vb{t}_{\lambda} = \lambda \vb{t}_0 + (1-\lambda)\vb{1}$ along the line of $\vb{t}_0$ and $\vb{1}$, their velocities are equal.
Therefore, it is easy to see such a constant velocity along the line connecting an off-diagonal point and the target point would lead to convergence to the target data point. Precisely,
\begin{align} 
\vb{x}_{\vb{t}_0} + \int_{t_0}^{1}\frac{\vb{x}_1 - \vb{x}_{\vb{t}_0}}{1-t_0}d{t} 
= \vb{x}_{\vb{t}_0} + \vb{x}_1 - \vb{x}_{\vb{t}_0} 
= \vb{x}_1,
\end{align}
due to the cone-shaped velocity field.
On the other hand, we note that the cone-shaped velocity introduces significant scaling by normalizing against the slowest time dimension, especially when the dimensions are scheduled very differently. 
This lack of regularity may cause the degraded regression of the velocity field, as demonstrated in later experiments.

\begin{table}[tb]
\centering
\caption{\FFF{\textbf{Ablation study on velocity adaptation methods.}
We compare no compensation, off diagonal drift minimization, and cone velocity on the first 100 ImageNet~\cite{ImageNet} categories. FID evaluates generation quality, while SSIM measures structural preservation under asynchronous generation with $\vb{t}_0=0.75$ in Fig.~\ref{fig:drift}.}}
\vspace{-3mm}
\resizebox{!}{0.04\textwidth}{
\begin{tabular}{c|ccc}
\toprule
Approaches  & w/o Cmpn & Off-diagonal Drift  & Cone Velocity  \\ \midrule
FID $\downarrow$ & 103.75   & \textbf{103.01} & 113.59 \\ 
SSIM $\uparrow$ in Fig.~\ref{fig:drift} & 0.255   & \textbf{0.262} & 0.224 \\ \bottomrule
\end{tabular}
}
\label{tab:drift}
\end{table}

\begin{figure}[!htb]
    \centering
    \begin{overpic}[width=0.95\linewidth]{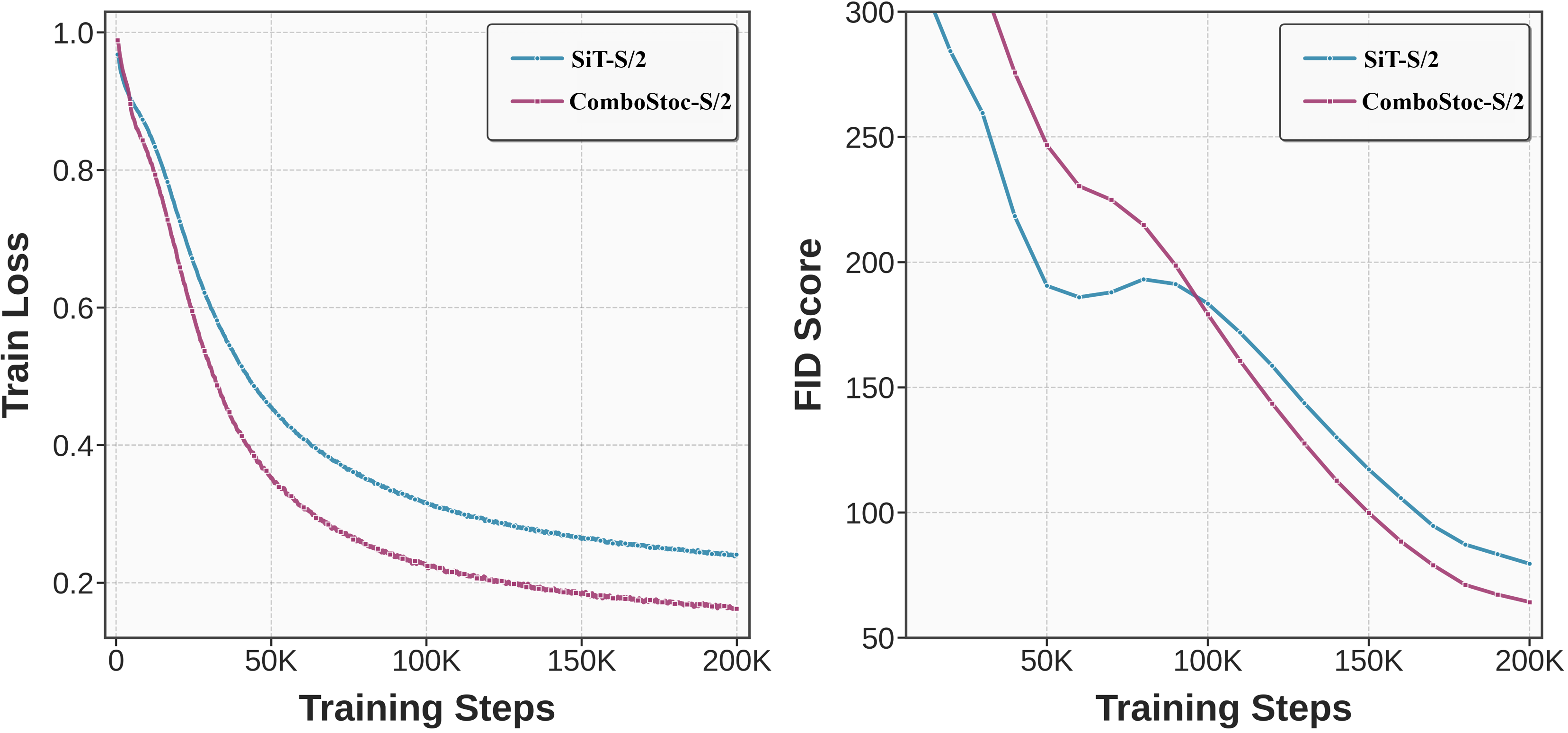}
    \end{overpic}
    \vspace{-2mm}
    \caption{\XR{\textbf{Training Loss and FID on Insufficient Data.}
    Training curves of SiT-S/2 (32.58M parameters) and ComboStoc-S/2 (32.36M parameters) on a small-scale dataset consisting of 1{,}000 images sampled from ImageNet~\cite{ImageNet}. Despite using slightly fewer parameters, ComboStoc consistently achieves a lower training loss and a better FID throughout training. FIDs are computed on images generated using ODE sampling with a classifier-free guidance scale of 1.0.}
    }
    \label{fig:1000loss}
\end{figure}
\paragraph{Ablation Study.}

We conducted ablation experiments on the two different velocity adaptation methods using a subset of ImageNet (the first 100 categories). The models were trained for 50,000 steps on 8 Nvidia H20 graphics cards. The first row in Tab.~\ref{tab:drift} presents the FID values obtained with these two methods, as well as without any compensation.
The results indicate that, compared to not using any compensation, our proposed off-diagonal drift minimization method achieves better convergence. In contrast, the cone-shaped velocity field results in poorer performance, which may be attributed to its scaled magnitude.
\FF{While the FID improvement from drift compensation appears modest in Tab.~\ref{tab:drift}, the visual impact is more pronounced: as shown in Fig.~\ref{fig:drift}, without compensation, images exhibit obvious background discontinuities along the boundary between differently preserved regions, whereas the off-diagonal drift minimization produces a seamless transition. These artifacts are visually salient but difficult to capture with scalar metrics alone.}

Additionally, similar to Fig.~\ref{fig:2dt0}, we examined the differences among three methods in asynchronous generation, as illustrated in Fig.~\ref{fig:drift}. 
We used different $\vb{t}_0$ values for each side of the image. 
The results show that the off-diagonal drift minimization method produces a more seamless transition. 
In contrast, without any compensation, a noticeable discontinuity appears along the midline of the image. 
The cone-shaped velocity field method, which converges more slowly, yields poorer results. 
It is important to note that these methods were trained for only 50,000 steps on a subset of ImageNet, so the overall image quality is relatively low.

To obtain a quantitative evaluation metric for the graded generation, we computed the structural similarity (SSIM) between each pair of reference image $\vb{x}_1$ and generated image using $\vb{t}_0=0.75$, for the three configurations. 
We have randomly selected 5K reference images for evaluation. 
The second row of Tab.~\ref{tab:drift} presents SSIM scores, indicating that the off-diagonal drift minimization method better preserved the structural similarity and quality to reference images, in comparison to the other settings.

Based on the above observation, we have used the off-diagonal drift minimization approach to mitigate the divergence issue in \XR{training and} all experiments.

\begin{table}[!t]
    \centering
    \caption{\FFF{\textbf{Comparison between uniform stepsize and uniform step number.}
We compare uniform stepsize (US; with different step numbers) and uniform step number (UN) for graded control under the same timestep setting as Fig.~\ref{fig:drift}, using our fully trained 800K model.}}
    \vspace{-2mm}
    \begin{tabular}{l|cc}
        \toprule
        Metric & US & UN \\
        \midrule
        SSIM $\uparrow$      & 0.175         & \textbf{0.188} \\
        PSNR $\uparrow$ & \textbf{8.124} & 8.050 \\
        LPIPS $\downarrow$   & \textbf{0.794} & 0.804 \\
        \bottomrule
    \end{tabular}
    \vspace{-2mm}
    \label{tab:us-vs-un}
\end{table}

\begin{figure*}[!tb]
    \centering
    \begin{overpic}[width=0.97\linewidth]{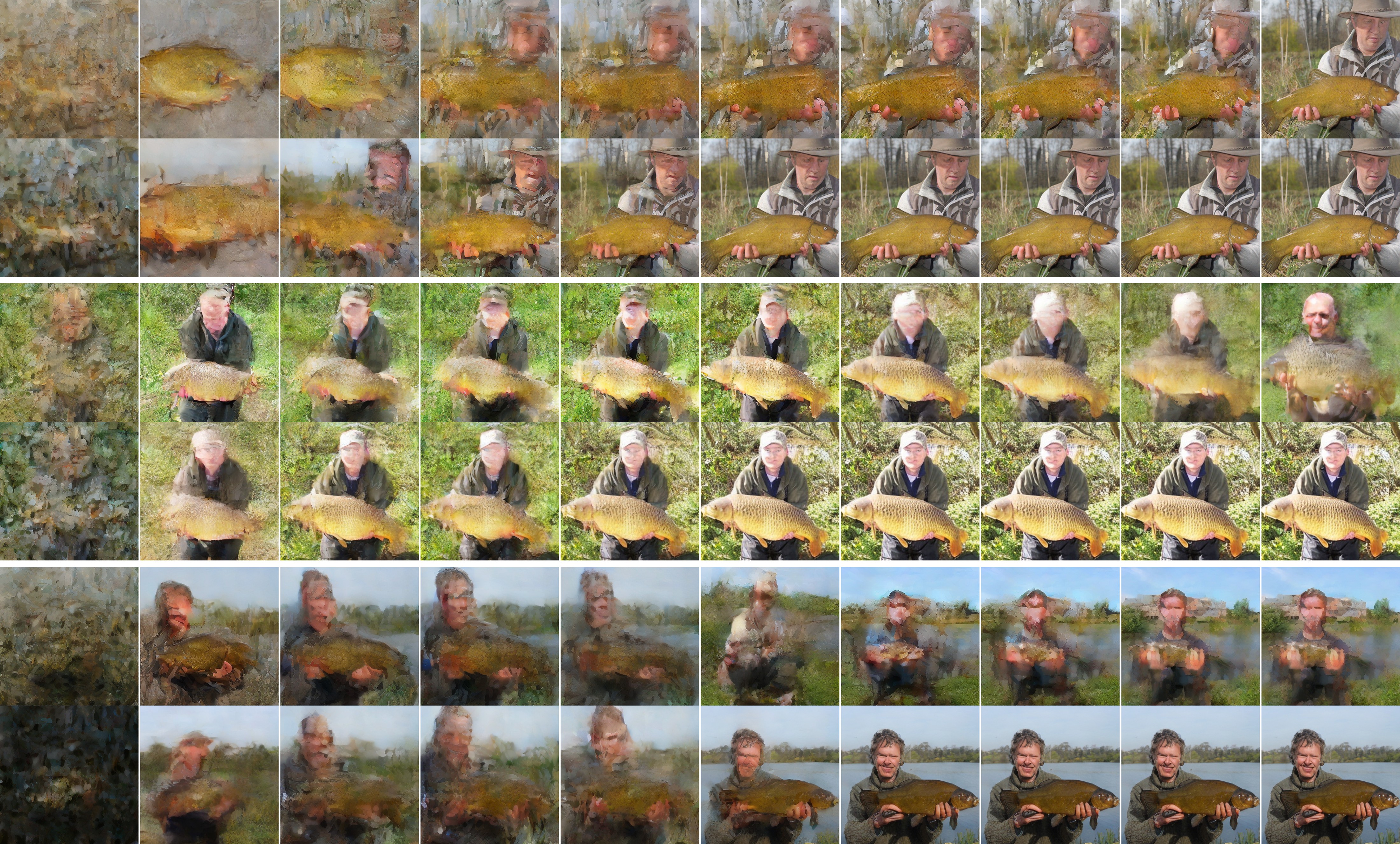}
    \put(3, -1.5)  {\small \textbf{20K}}
    \put(14,-1.5)  {\small \textbf{40K}}
    \put(24,-1.5)  {\small \textbf{60K}}
    \put(34,-1.5)  {\small \textbf{80K}}
    \put(44,-1.5)  {\small \textbf{100K}}
    \put(53.5,-1.5){\small \textbf{120K}}
    \put(63.5,-1.5){\small \textbf{140K}}
    \put(73.5,-1.5){\small \textbf{160K}}
    \put(83.2,-1.5){\small \textbf{180K}}
    \put(93.5,-1.5){\small \textbf{200K}}
    \put(-1.4,1){\rotatebox{90}{\small  \textbf{ComboStoc}}}
    \put(-1.4,14){\rotatebox{90}{\small \textbf{SiT}}}
    \put(-1.4,21){\rotatebox{90}{\small \textbf{ComboStoc}}}
    \put(-1.4,34){\rotatebox{90}{\small \textbf{SiT}}}
    \put(-1.4,42){\rotatebox{90}{\small \textbf{ComboStoc}}}
    \put(-1.4,54){\rotatebox{90}{\small \textbf{SiT}}}
    \end{overpic}
    \caption{\XR{\textbf{Visual Comparison with SiT-S/2 under Insufficient Data.}
    Qualitative results of SiT-S/2 (32.58M) and ComboStoc-S/2 (32.36M) at different training iterations.
    Both models are trained and evaluated using the same random seed, with inference performed every 20K training steps. The images are generated using ODE sampling with a classifier-free guidance scale of 1.0.}
    }
    \label{fig:1000image}
\end{figure*}
\subsubsection{Async timestep at inference stage.}
\label{sec:step}
\XR{
For the graded control applications on both image and 3D shape, currently we use different stepsizes but the same step number \(N\) for individual dimensions.
Therefore, regions of larger \(t_{0}\) will go through the same \(N\) iterations as those of smaller \(t_{0}\) in all of the applications. This scheme differs from the formula in Sec.~\ref{sec:proof}, where a uniform stepsize is applied to analyze the drift problem during training.

Meanwhile, we tested the uniform stepsize (different step number, \textbf{US}) scheme for graded control in Tab.~\ref{tab:us-vs-un} under the initial timestep setting in Fig.~\ref{fig:drift} but using our fully trained 800K steps model (not the 50,000 steps subset used in Fig.~\ref{fig:drift}), and found the results are similar to the uniform step number (\textbf{UN}) scheme. 


}

\subsubsection{\XR{Ablation on Insufficient Data}}
\label{subsubsec:imagenet_oneclass}
\XR{
To further evaluate the effectiveness of \name \space under data-scarce regimes, we conduct an ablation study on insufficient training data.
We adopt the smallest SiT-S/2 model (32.58M parameters) as the baseline and construct ComboStoc-S/2 on top of it, which has slightly fewer parameters (32.36M) due to different time embedding module.
Both models are trained from scratch on a small subset of ImageNet~\cite{ImageNet} containing only 1{,}000 images in one category.}

\XR{
Fig.~\ref{fig:1000loss} presents the quantitative comparison between SiT-S/2 and ComboStoc-S/2 on the insufficient dataset.
Although ComboStoc-S/2 contains slightly fewer parameters, it consistently achieves a lower training loss than SiT-S/2 throughout almost the entire training process.
In terms of FID, SiT-S/2 performs slightly better at early stages, when the learned representations remain coarse.
As training proceeds and more meaningful structure emerges, ComboStoc-S/2 progressively closes the gap and ultimately outperforms SiT-S/2, leading to a clear advantage in the later stages.
This behavior indicates that once meaningful structure begins to emerge, \name \space is able to exploit structured information more effectively, leading to richer representations and improved generation quality under limited data.

Fig.~\ref{fig:1000image} further illustrates this advantage through visual comparisons across training iterations.
Using identical random seeds for both training and inference, we observe that \name \space produces higher-quality and more stable images at significantly earlier training stages.
In contrast, SiT often remains blurry even after 200K steps in several cases, or exhibits chaotic outputs in early training and only converges to clear images after extensive optimization.
Overall, \name \space demonstrates faster convergence and stronger robustness in low-data regimes, consistently yielding clear and semantically coherent samples midway through training.
}

\begin{table}[tp!]
\centering
\caption{\FFF{\textbf{Comparison of computational complexity with SiT and DiT.}
We compare parameter count, training speed and memory usage, inference speed, and GFLOPs. All evaluations are conducted on a single NVIDIA A100 80GB GPU using the XL/2 model configuration with 256$\times$256 input images. GFLOPs are computed using DeepSpeed.}}
\vspace{-2mm}
\resizebox{!}{0.071\linewidth}{
\begin{tabular}{l|c|c|c|c|c}
\toprule
Methods & Parameters(M) & Mem. Usage(MB) & Training Speed (steps/sec) & GFlops & Inference Speed(ms) \\ \midrule
DiT                    & 675           & 75580          & 0.17                       & 237.34 & 49                  \\ \hline
SiT                    & 675           & 76868          & 0.19                       & 237.34 & 50                  \\ \hline
Ours                   & 673           & 76340          & 0.15                       & 352.46 & 48                  \\ \bottomrule
\end{tabular}
}
\label{tab:complexity}
\end{table}
\subsection{Computational Complexity Analysis}
\label{sec:app_complexity}

We provide a comparison with SiT and DiT in Tab.~\ref{tab:complexity}, in terms of parameter count, training stage speed, memory usage, inference stage speed, and GFlops. All tests are done on a single Nvidia A100-80G GPU at the XL/2 model configuration, with an input image of size 256$\times$256 and training batch size 256. The GFlops are calculated by DeepSpeed~\citep{deepspeed}.

From Tab.~\ref{tab:complexity} we can see that compared with DiT and SiT, our model has a smaller number of parameters as we use a smaller timestep embedding module (see Fig.~\ref{fig:emb}). Therefore, our GPU memory cost during training is slightly smaller than SiT. On the other hand, for the conditioning by class label and timestep implemented as a modulation operator (see Fig. 3 of the DiT paper for illustration~\citep{DiT_2023_ICCV}), our conditioning is a tensor of the same shape as the image tensor, in contrast to DiT/SiT’s conditioning by a vector only of the channel size of the image tensor; to produce the conditioning tensor involves more computation than the conditioning vector, so our model leads to more flops and slightly increased training cost per step. Nevertheless, the production of the conditioning tensor is a standard MLP feature transformation and fits nicely into GPU parallel computation, so the inference speed is not sacrificed in comparison with DiT/SiT.
\FF{In summary, the per-step training slowdown ($0.15$ vs.\ $0.19$ steps/sec for SiT) is a direct consequence of the tensorized timestep conditioning; however, as shown in Fig.~\ref{fig:2dfid} (c), \name\ still achieves better FID at the same wall-clock training time, indicating that the quality gains more than compensate for the per-step overhead.}

\section{Conclusion}
\label{sec:conclusion}

We have proposed to focus on the problem of combinatorial complexity of high-dimensional and multi-attribute data samples (like 3D shapes and images) for diffusion generative models.
In particular, we note that for one-sided stochastic interpolants that model many variants of diffusion and flow-based models, there exists the problem of under-sampling regions of the path space where the dimensions/attributes are off-diagonal or asynchronous.
We propose to fix this issue by sampling the whole space spanned by combinatorial complexity uniformly.
Experiments across two data modalities show that indeed by utilizing the combinatorial complexity, performances can be enhanced, and new generation paradigms can be enabled where different attributes of a data sample are generated in asynchronous schedules to achieve varying degrees of control simultaneously.
We hope that our work can inspire future works that look through the combinatorial perspective of generative models.

\paragraph{Limitations and future work.}
Our \name \space scheme will only have significant effects when the data has combinatorial and structural information, such as different patches for images and different parts for 3D shapes. When the data resides in a vector space whose dimensions are nearly independent, it is hard to exploit the correlation of dimensions and train a model that works well under the combinatorial schedule of different dimensions.
Indeed, in such a case the individual dimensions may ideally be generated separately by different models.
However, we note that many data types in real life contain strong structural and combinatorial information; particularly eminent are tasks within scientific domains, including molecule docking and protein folding~\citep{corso2022diffdock, wu2024protein, yim2024diffusion}, where diffusion models that better handle their combinatorial structures can be desirable. Additionally, \name \space is not specifically designed for diffusion transformers. Instead, it can be easily applied to other networks like U-Net diffusers; we only need to tensorize the timestep and apply it to corresponding feature maps through elementwise modulation, where the modulation can be the standard conditioning (e.g. the affine transformation).


\bibliographystyle{ACM-Reference-Format}
\bibliography{src/comboStochastic}

\clearpage

\end{document}